\documentclass[10pt,twocolumn,letterpaper]{article}

\usepackage{iccv}
\usepackage{times}
\usepackage{epsfig}
\usepackage{graphicx}
\usepackage{amsmath}
\usepackage{amssymb}
\usepackage{booktabs}
\usepackage{multirow}
\usepackage{bm}
\usepackage{adjustbox}
\usepackage{pifont}
\usepackage{diagbox}
\usepackage{xcolor}
\usepackage{mathtools}
\usepackage{algorithm}
\usepackage{subcaption}
\usepackage{cite}

\usepackage[pagebackref=true,breaklinks=true,letterpaper=true,colorlinks,bookmarks=false]{hyperref}

\iccvfinalcopy 

\def\iccvPaperID{****} 

\ificcvfinal\fi

\begin{document}

\title{BlendFace: Re-designing Identity Encoders for Face-Swapping}

\author{Kaede Shiohara$^{1}$\thanks{Work done during an internship at CyberAgent AI Lab} \quad Xingchao Yang$^{2}$ \quad Takafumi Taketomi$^{2}$ \qquad \vspace{1pt}\\
$^{1}$The University of Tokyo  \qquad $^{2}$CyberAgent AI Lab\qquad\qquad\\
\hspace{0.1in}{\tt\small shiohara@cvm.t.u-tokyo.ac.jp} \qquad  {\tt\small \{you\_koutyo,taketomi\_takafumi\}@cyberagent.co.jp} \\
}

\maketitle
\ificcvfinal\thispagestyle{empty}\fi

\begin{abstract}
The great advancements of generative adversarial networks and face recognition models in computer vision have made it possible to swap identities on images from single sources. 
Although a lot of studies seems to have proposed almost satisfactory solutions, we notice previous methods still suffer from an identity-attribute entanglement that causes undesired attributes swapping because widely used identity encoders, \eg, ArcFace, have some crucial attribute biases owing to their pretraining on face recognition tasks.
To address this issue, we design BlendFace, a novel identity encoder for face-swapping. The key idea behind BlendFace is training face recognition models on blended images whose attributes are replaced with those of another mitigates inter-personal biases such as hairsyles.
BlendFace feeds disentangled identity features into generators and guides generators properly as an identity loss function. 
Extensive experiments demonstrate that BlendFace improves the identity-attribute disentanglement in face-swapping models, maintaining a comparable quantitative performance to previous methods.
The code and models are available at \url{https://github.com/mapooon/BlendFace}.
\end{abstract}

\section{Introduction}

\begin{figure}[t]
    \centering
    \includegraphics[width=0.99\linewidth]{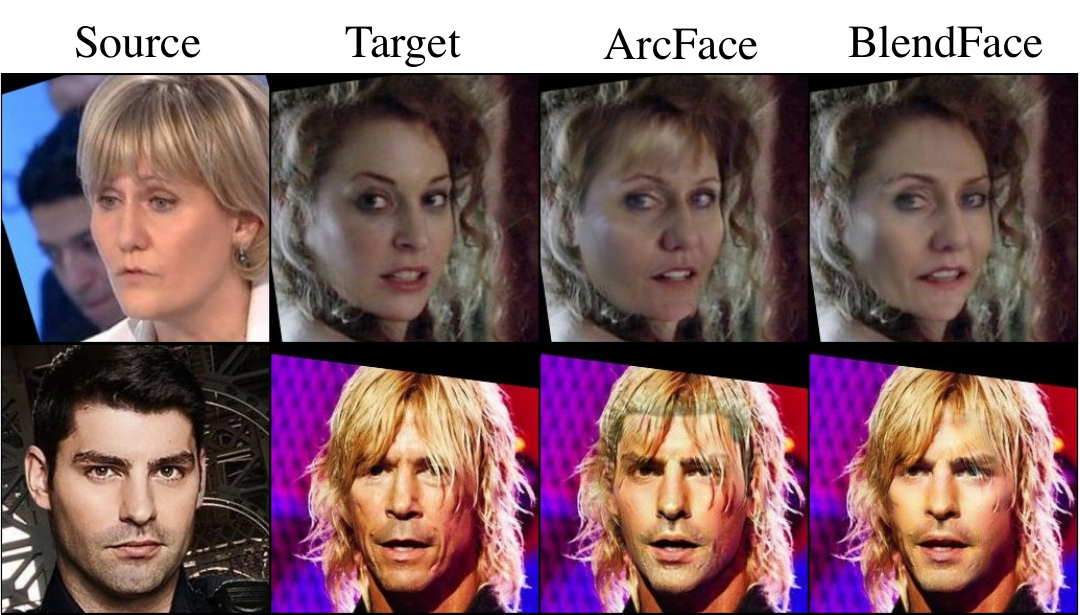}
    \caption{\textbf{Examples of attribute leakages.} The direct use of face recognition models, \ie, ArcFace~\cite{arcface} as an identity guidance causes attribute leakages, especially on head shapes and hairstyles due to attribute biases inherent in the face recognition models whereas our identity encoder BlendFace address these problems. Best viewed in zoom.}
    \label{teaser}
\end{figure}

Face-swapping aims to replace target identities with source identities in images while preserving the target attributes, \eg, facial expression, hair, pose, and background. This task is receiving considerable attention because of its potential applications in various fields, such as films and metaverses.
Recent advances in generative adversarial networks (GANs)~\cite{gan,dcgan,lsgan,wgan,CycleGAN2017,progan,stylegan} have enabled the photo-realistic image generation in various conditions, \eg, attribute~\cite{attgan}, identity~\cite{ipgan}, and expression~\cite{reenact}, as well as unconditional image generation.
Moreover, the advancement in face recognition models provides powerful identity encoders for face-swapping, which boosts the transferability of identities from source inputs to generated images and leads to successful one-shot face-swapping models~\cite{ipgan,simswap,faceshifter,hififace,megafs,infoswap,rafswap,fslsd,styleswap,styleface} with reasonable quality.

However, despite these impressive efforts, a critical issue still remains.
Previous state-of-the-art methods suffer from identity-attribute entanglements because of biased guidance from face recognition models used as identity encoders. Fig.~\ref{teaser} presents the failure cases of a traditional face recognition model ArcFace~\cite{arcface}. As shown in the figure, ArcFace-based face-swapping models tend to swap undesired attributes, \eg, hairstyles and head shapes. 
This is because images of the same identity have strong correlations for some attributes; therefore, face recognition models accidentally learn to recognize the attributes as identities, which causes misguidances in training face-swapping models.
Though certain studies in the field of face recognition propose effective approaches to mitigate biases between individuals, they cannot be solutions for biases in face-swapping models as they do not consider intra-personal biases.

\begin{figure}[t]
    \centering
    \includegraphics[width=0.99\linewidth]{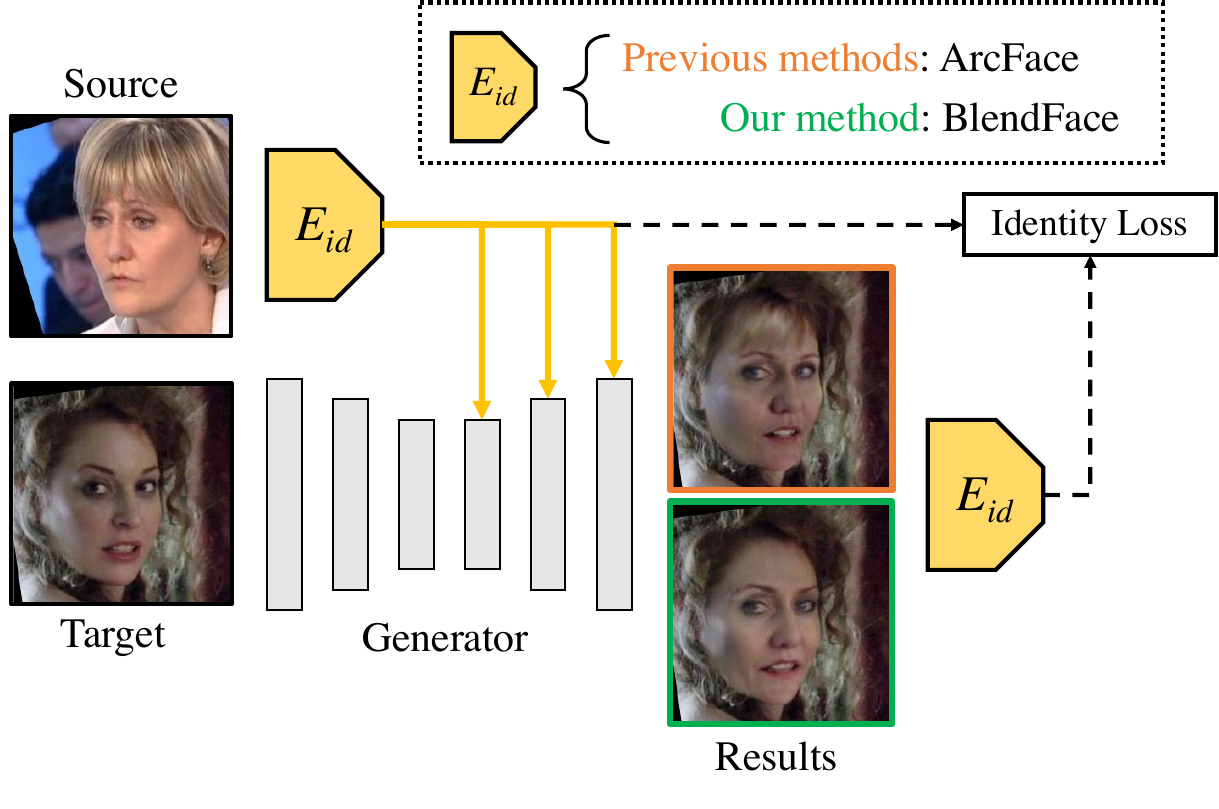}
    \caption{\textbf{Overview of face-swapping models.} Identity encoder $E_{id}$ is used to extract identity features from source images and to guide generators as an identity distance loss. We replace ArcFace with our identity encoder BlendFace to achieve more disentangled face-swapping.}
    \label{overview}
\end{figure}

In this paper, we propose BlendFace, a novel identity encoder that provides well-disentangled identity features for face-swapping. 
First, we analyze the widely used identity encoder ArcFace~\cite{arcface} on VGGFace2~\cite{vggface2} dataset. 
The comparison of identity similarity distributions using pseudo-swapped images clarifies that attributes such as hairstyles, colors, and head shapes strongly affect their similarities because of attribute biases in ArcFace, which is expected to prevent face-swapping models from disentangling identity and attribute. 
Based on the observation from the preliminary experiment, we design BlendFace by simply training ArcFace with swapped images so that the model does not focus on attributes of faces, which bridges the gap between similarity distributions of swapped faces and real faces. 
We then train a face-swapping model using BlendFace that performs as a source feature extractor and identity guidance in the loss function. 
As shown in Fig.~\ref{overview}, by replacing the traditional identity encoder in the source feature extraction and loss computation with BlendFace, face-swapping models are trained to generate more disentangled face-swapping results. 
Importantly, our work is compatible with previous face-swapping studies; BlendFace can be applied to various learning-based face-swapping models.

In the experiment part, we compare our model with state-of-the-art face-swapping models on FaceForensics++~\cite{ffpp} dataset following the convention.
The comparison demonstrates that the proposed method is superior to or on par with the competitors in identity similarity and attribute preservation, \ie, expression, pose, and gaze, while improving visual consistency of swapped results compared to previous models.
In addition, our ablation study and analysis prove the advantages of BlendFace from various perspectives for face-related research.

\section{Related Work}
\noindent \textbf{Face-Swapping.} Face manipulations, particularly face-swapping, is an area of significant research in computer vision owing to its potential applications such as realistic digital avatar creation. 
Early methods employ traditional image processing~\cite{10.1145/1360612.1360638} and 3D morphable models (3DMMs) ~\cite{Blanz2004ExchangingFI,10.1145/1599301.1599330,nirkin2018face}.  
The brilliant successes of generative adversarial networks (GANs)~\cite{gan,dcgan,lsgan,wgan,CycleGAN2017,progan,stylegan} in computer vision has driven the extensive exploration of learning-based face-swapping models.
FSGAN~\cite{fsgan} realizes subject-agnostic face-swapping via four encoder-decoder networks, \textit{i.e.}, reenactment, segmentation, inpainting, and blending networks.
Sophisticated face recognition models~\cite{arcface,cosface,adaface,sphereface} that learn rich identity information from large-scale facial datasets~\cite{cisia,webface260m,vggface2,ms1m,glint360k} improves the identity preservation in face swapping.
SimSwap~\cite{simswap} proposes a weak feature matching loss between each generated image and target image in the discriminator's feature space to balance the preservation of the source identity and target attribute.
FaceShifter~\cite{faceshifter} proposes a two-stage framework including AEI-Net that blends the features extracted from source and target images in multiple scales and HEAR-Net that learns to reconstruct occlusions using objects datasets~\cite{egohands,gtea,shapenet}.
InfoSwap~\cite{infoswap} introduces information-theoretical loss functions to disentangle identities.
HifiFace~\cite{hififace} incorporates a 3DMM~\cite{deep3dfacerecon} to its identity extraction to retrain the source appearance and shape. 
Smooth-Swap~\cite{smoothswap} develops a smooth identity encoder to stably GAN-training using self-supervised pretraining~\cite{scl}.
Recently, some studies~\cite{megafs,rafswap,fslsd} reveal pretrained StyleGANs~\cite{stylegan,stylegan2,stylegan3} provide strong priors to generate photo-realistic facial images at megapixel resolution for face-swapping.
MegaFS~\cite{megafs} generates swapped faces by replacing high semantic features of target images with those of source images.
RAFSwap~\cite{rafswap} integrates semantic-level features with a face-parsing model~\cite{faceparsing}.
FSLSD~\cite{fslsd} transfers multi-level attributes via side-outputs from StyleGAN.
StyleSwap~\cite{styleswap} proposes iterative identity optimization that effectively preserves source identities.
StyleFace~\cite{styleface} and UniFace~\cite{uniface} unify face-swapping into de-identification and reenactment, respectively.
In this paper, we re-design identity encoders independently of these state-of-the-art approaches; our encoder can be easily incorporated into previous learning-based face-swapping models.

\vskip.5\baselineskip
\noindent \textbf{Face Recognition.} The task of face recognition is a fundamental problem in the research field. Recent approaches have mainly been conducted using deep convolutional networks. In particular, margin-based loss functions (\eg, \cite{arcface,cosface,sphereface}) significantly enhance the performance of face recognition.
However, some studies~\cite{terhorst2021comprehensive,robinson2021balancing,yucer2020exploring,9025674,facegenderid,issuesrelated} have found such identity encoders contain biases of attributes, \eg, pose, hairstyles, color, races, and gender; therefore debiasing face recognition models have been a concern topic in the field.
IMAN~\cite{iman} proposes information-theoretical adaptation networks. 
RL-RBN~\cite{rlrbn} adapts the margin of ArcFace by reinforcement learning.
GAC~\cite{gac} proposes adaptive layers comprising demographic-group-specific kernels and attention modules.
DebFace~\cite{debface} disentangles gender, age, race, and identity using feature disentangling blocks and aggregation blocks in an adversarial learning.

Although these methods effectively mitigate the biases between identities, they however do not focus on inter-personal biases. 
Therefore, existing identity extractors cause undesired attributes swapping because images of each identity in datasets used for face recognition have strong correlations in some attributes, \eg, hairstyles, colors, and head shapes. 
To solve this problem, we design an debiased encoder that extracts disentangled identity features from facial images by training a face recognition model with synthetic images that have swapped attributes, which enables well-disentangled face-swapping.

\begin{figure}[t]
    \centering
    \includegraphics[width=0.99\linewidth]{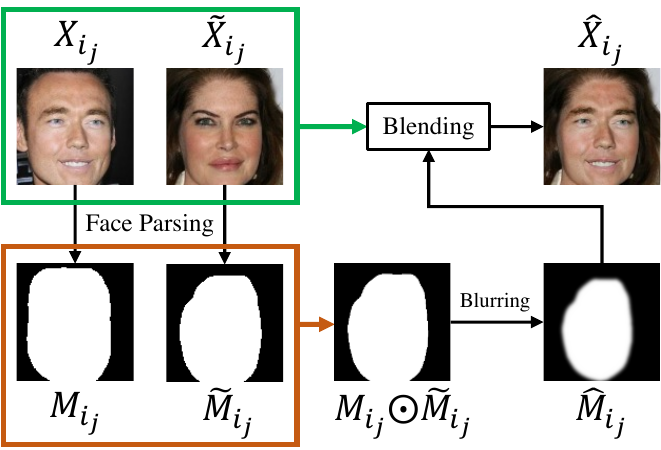}
    \caption{\textbf{The pipeline for pseudo-positive samples.} We first generate masks ${M}_{i_j}$ and $\tilde{M}_{i_j}$ corresponding to $X_{i_j}$ and $\tilde{X}_{i_j}$, respectively, then generate a smooth mask $\hat{M}_{i_j}$ by blurring the intersection mask $M_{i_j} \odot \hat{M}_{i_j}$. Finally we blend $X_{i_j}$ and $\hat{X}_{i_j}$ with $\hat{M}_{i_j}$ to generate the blended image $\hat{X}_{i_j}$.}
    \label{fig:blending}
\end{figure}

\section{Attribute Bias in Face Recognition Models} 
\label{sec:attribute_bias}

Given source and target images, face-swapping aims to generate a facial image where the target identity is replaced with the source identity while preserving the attributes of the target image. 
First of all, we rethink the identity encoding for face-swapping by conducting a preliminary experiment with ArcFace~\cite{arcface} adopted by most face-swapping models, \eg, \cite{simswap,faceshifter,megafs,fslsd,hififace,infoswap}. 
The key observation is replacing attributes of one individual with those of another causes a degradation of the identity similarity, which indicates attribute biases inherent in ArcFace.

\subsection{Identity Distance Loss}
One of the difficulties in face-swapping is the absence of ground truth images. 
Given two images of different identities for source and target inputs during the training, generated images are constrained by some feature-based losses, \eg, appearance~\cite{lpips,arcface,perceptual}, 3D face shapes~\cite{deep3dfacerecon}, and segmentation~\cite{faceparsing} to preserve identities of source and attributes of target. 
Notably, most of the previous methods adopt ArcFace~\cite{arcface} trained on large-scale face recognition datasets~\cite{cisia,webface260m,vggface2,ms1m,glint360k} to extract identity information from source inputs and measure the identity distance between a source image $X_s$ and swapped image $Y_{s,t}$ as follows:
\begin{equation}
\label{traditional_id_loss}
    \mathcal{L}_{id}=1-cos\langle E_{id}(X_{s}), E_{id}(Y_{s,t})\rangle,
\end{equation}
where $E_{id}$ denotes ArcFace encoder and $cos\langle u,v\rangle$ is the cosine similarity of vectors $u$ and $v$.

\begin{figure}[t]
    \centering
    \begin{minipage}[b]{0.62\linewidth}
    \centering
    \includegraphics[width=0.99\linewidth]{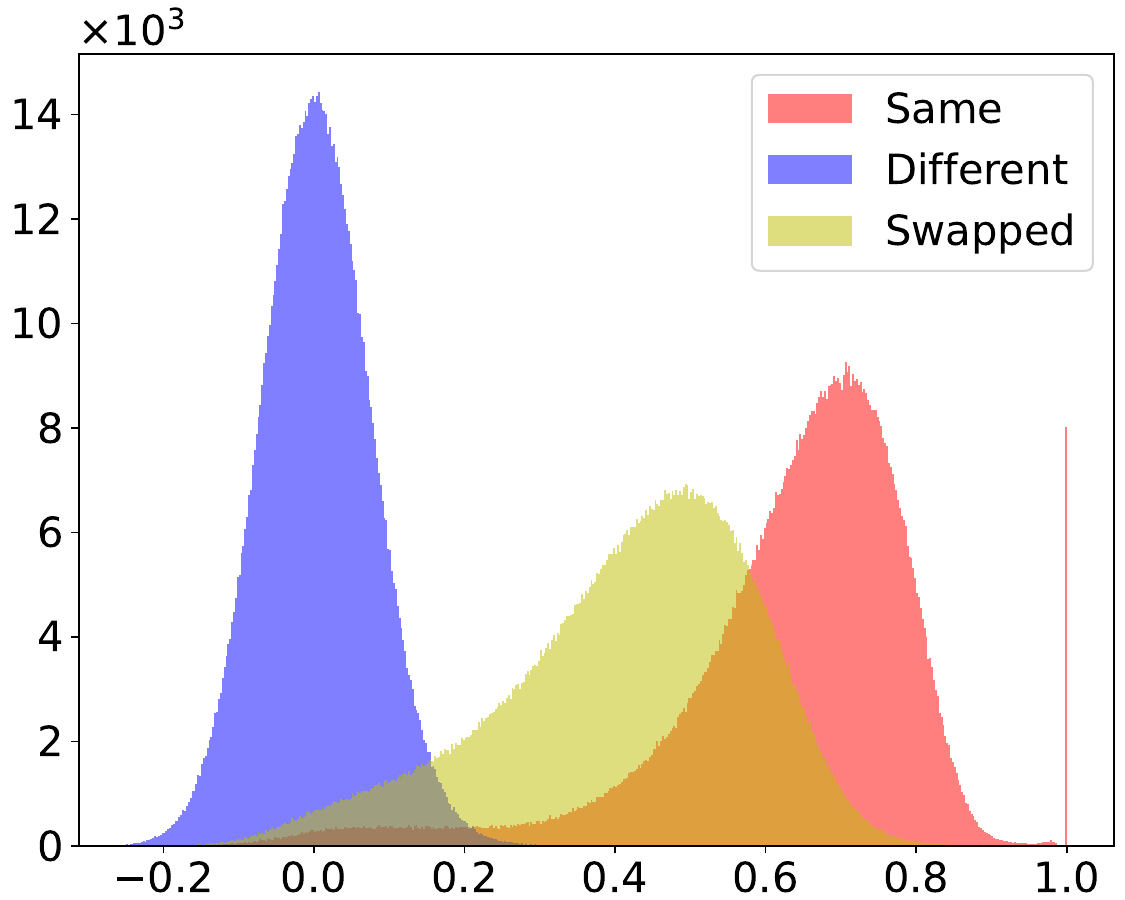}
    \subcaption{Similarity distributions}
    \label{preliminary_distribution}
    \end{minipage}
    \begin{minipage}[b]{0.37\linewidth}
    \centering
    \includegraphics[width=0.99\linewidth]{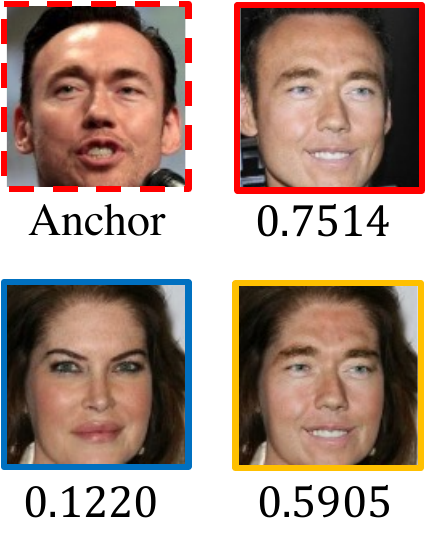}
    \subcaption{Examples}
    \label{preliminary_sample}
    \end{minipage}
    \caption{\textbf{Analysis of ArcFace on VGGFace2.} ArcFace tends to underestimate the identity similarities between anchor images and swapped faces.}
    \label{preliminary}
\end{figure}

\subsection{Analysis of Identity Similarity}
\label{sec:preliminary}

Here, we explore the attribute biases of ArcFace~\cite{arcface} on VGGFace2~\cite{vggface2} dataset from a perspective of face-swapping. 
As shown in Fig.~\ref{fig:blending}, we first randomly sample the $j$-th image of the $i$-th identity  which is presented as $X_{i_j}$, and then  compute cosine similarities between $X_{i_j}$ and all the images of the same identity $\{X_{i_1}, X_{i_2}, \cdots, X_{i_{n_i}}\}$, where $n_i$ denotes the number of images of the identity $i$. 
Subsequently, inspired by Face X-ray~\cite{facexray}, we search for an image $\tilde{X}_{i_j}$ with the closest facial landmarks to $X_{i_j}$ from randomly sampled 100 images whose identities are not $i$ for each $X_{i_j}$. After transferring the color statistics $\mu$ and $\sigma$ in the Lab space of $X_{i_j}$ to that of $\tilde{X}_{i_j}$, we replace the face of $\tilde{X}_{i_j}$ with that of $X_{i_j}$ by blending them with a mask $\hat{M}_{i_j}$ generated by multiplying the inner mask $M_{i_j}$ of $X_{i_j}$ by $\tilde{M}_{i_j}$ of $\tilde{X}_{i_j}$:
\begin{equation}
\hat{X}_{i_j}= X_{i_j} \odot \hat{M}_{i_j} + \tilde{X}_{i_j} \odot (1-\hat{M}_{i_j}),
\end{equation}
where $\odot$ denotes the point-wise product, $\hat{X}_{i_j}$ represents the synthetic swapped image, and $\hat{M}_{i_j} = Blur(M_{i_j} \odot \tilde{M}_{i_j})$. 
We then compute the cosine similarities between $X_{i_j}$ and replaced images $\{\hat{X}_{i_1}, \hat{X}_{i_2}, \cdots, \hat{X}_{i_{n_i}}\}$.
In addition, we calculate cosine similarities between $X_{i_j}$ and the closest images $\tilde{X}_{i_j}$.
We repeat this procedure for all identities and plot the similarity distributions in Fig.~\ref{preliminary_distribution}. 

The key observations of this experiment are as follow:
1) The similarities are equal to one only for identical image comparisons; otherwise, the similarities are almost lower than $0.85$ even if two images have the same identity. 
2) The similarities between anchor images and synthetic positive images are lower than those between actual positive pairs, indicating that the color distribution of faces and attributes of outer face region strongly affect the similarities. 
This is because face recognition models tend to recognize certain attributes, \eg, hairstyles and head shapes as identities because an image set of each identity of face recognition datasets used for training often has correlations in their attributes. 

These results leads us to assume that minimizing the identity loss in Eq.~\ref{traditional_id_loss} using traditional face recognition models, \eg, ArcFace~\cite{arcface} conflicts with the presevation of target attributes because face recognition models have attribute biases, which constrains generators excessively to transfer not only identities but also attributes from source images.

\section{BlendFace}
We propose a novel identity encoder BlendFace to solve the problem of the identity-attribute entanglement owing to the attribute biases in face recognition models as discussed in Sec.~\ref{sec:preliminary}.
First, we introduce a pre-training strategy to train debiased identity encoders using synthetically swapped faces. We then incorporate our identity encoder into a face-swapping model disentangling attributes and identities for high-fidelity face-swapping.

\subsection{Pre-training with Swapped Faces}
As discussed in Sec.~\ref{sec:preliminary}, traditional face recognition models trained on real face datasets, \eg, MS-Celeb-1M~\cite{ms1m} learn accidentally attribute biases because images of each identity are highly correlated with each other in some attributes, \eg, hairstyles and makeup; this produces poor results for source and target images with large attribute differences. 
To tackle this problem, we develop a debiased identity encoder BlendFace that can be achieved by training a face recognition model with synthetic facial images whose attributes are swapped.
We adopt ArcFace~\cite{arcface} as our base model and train it with blended images that have synthetically swapped attributes.
For each sample during the training, we swap attributes of input images in the same manner as in Sec.~\ref{sec:preliminary} with probability $p$. 
The loss function of ArcFace~\cite{arcface} is as follows:
\begin{equation}
\label{loss_pretraining}
\mathcal{L} = - \log \frac{e^{s\cos(\theta_{y_i} + m)}}{e^{s\cos(\theta_{y_i} + m)}+\sum_{k=1, k\ne y_i}^{K}e^{s\cos\theta_k}},
\end{equation}
where $\theta_{y_i}$ represents the angle between the deep feature vector and weight vector of the encoder. $K$, $s$, and $m$ denote the number of classes, scale, and margin, respectively.
We observe that our pretraining bridges the gap between the distribution of ``Swapped'' and that of ``Same'' (see Fig.~\ref{hist_p050}).
We conduct ablations for $p$ and $\hat{M}_{i_j}$ in Sec.~\ref{sec:ablation}.

\subsection{Face-Swapping with BlendFace}
To validate the effectiveness of BlendFace, we construct a face-swapping model with BlendFace. 
We denote source, target, 
 and generated images as $X_s$, $X_t$, and $Y_{s,t}$ $(=G(X_s, X_t))$, respectively. 
We adopt a state-of-the-art architecture AEI-Net~\cite{faceshifter} with some modifications. 
We replace ArcFace used in encoding source identities and computing the distance loss $\mathcal{L}_{id}$ (Eq.~\ref{traditional_id_loss}) with BlendFace.
We incorporate a blending mask predictor into the attributes encoder inspired by previous studies (\eg, \cite{hififace,rafswap,styleswap}). A predicted mask $\hat{M}$ is supervised by the binary cross entropy loss $\mathcal{L}_{mask}$ with the ground truth mask $M$ from a face-parsing model~\cite{faceparsing} as follows:
\begin{equation}
\mathcal{L}_{mask}= -\sum_{x,y}\{M_{x,y}\log\hat{M}_{x,y} + (1-M_{x,y})\log(1-\hat{M}_{x,y})\},
\end{equation}
where $x$ and $y$ are the spacial coordinates of image.
We feed different images that share the same identity for source and target inputs rather than feeding the same image when activating the reconstruction loss as follows:
\begin{equation}
  \mathcal{L}_{rec}=
  \begin{cases}
    \left\|X_{t} - Y_{s,t}\right\|_1 & \text{if $ID(X_t) = ID(X_s)$,} \\
    0      & \text{otherwise.}
  \end{cases}
\end{equation}
We sample the same identity for source and target images with $p=0.2$.
We use the cycle consistency loss instead of the attributes loss used in the original FaceShifter as follows:
\begin{equation}
\mathcal{L}_{cyc}= \left \| X_t- G(X_t, Y_{s,t})\right\|_1.
\end{equation}
We use the same adversarial loss term $\mathcal{L}_{adv}$ as in GauGAN~\cite{gaugan}.
The total loss $\mathcal{L}$ is formulated as:
\begin{equation}
\label{total_loss}
\mathcal{L}= \mathcal{L}_{adv} + \mathcal{L}_{mask} + \lambda_{1}\mathcal{L}_{id} +  \lambda_{2}\mathcal{L}_{rec} + \lambda_{3}\mathcal{L}_{cyc},
\end{equation}
where the coefficients $\lambda_{1}$, $\lambda_{2}$, and $\lambda_{3}$ are hyper-parameters that balance the loss functions.
Please see the supplementary material for a more detailed description of the architecture of our face-swapping model.

\begin{table*}[t]
    \centering
\resizebox{\textwidth}{!}{
{\small
\begin{tabular}{lcccccccccccc}
\toprule

\multirow{2}{*}{Model} & \multicolumn{6}{c}{Identity Distance} & \multicolumn{6}{c}{Attribute Distance}\\
\cmidrule(lr){2-7} \cmidrule(lr){8-13}
&      Arc &   Arc-R &      Blend &   Blend-R &   Shape &   Shape-R &   Expr &   Expr-R &   Pose &   Pose-R & Gaze & Gaze-R \\
\midrule
Deepfakes\cite{deepfake-faceswap}$^*$ &\underline{0.5388} & \underline{0.4048} & 0.3704 & 0.2859 & 0.3339 & 0.4327 & 0.2037 & - & 0.0374 & - & 0.2891 & -\tabularnewline
FaceSwap~\cite{faceswap}$^*$ & 0.5916 & 0.4594 & 0.4653 & 0.3606 & 0.3355 & 0.4546 & 0.1824 & - & 0.0189 & -& 0.2273 & -\tabularnewline
FSGAN~\cite{fsgan} & 0.7188 & 0.5043 & 0.6640 & 0.4464& 0.3918 & 0.5410 & \textbf{0.1419} & \textbf{0.4212} & 0.0173 & 0.1835 & 0.1772& 0.4120 \tabularnewline
FaceShifter \cite{faceshifter}$^*$ &  {\color[gray]{0.7}0.3826} & {\color[gray]{0.7}0.3197} & \textbf{0.3143} & \textbf{0.2607} & 0.3190 & 0.4516 & 0.1720 & - & 0.0162 & - & 0.1840 & -\tabularnewline
SimSwap \cite{simswap}& {\color[gray]{0.7}0.3594} & {\color[gray]{0.7}0.3012} & 0.3710 & 0.2992 & 0.3309 & 0.4761 & 0.1582 & 0.4809 & \underline{0.0139} & \underline{0.1512} & \underline{0.1599} & \underline{0.4052}\tabularnewline
HifiFace \cite{hififace}$^*$ & {\color[gray]{0.7}0.3593} & {\color[gray]{0.7}0.2940} & \underline{0.3488} & \underline{0.2771}& \underline{0.3119} & 0.4405 & 0.1730 & 0.5066 & 0.0161 & 0.1702 & 0.1663 & 0.4183\tabularnewline
MegaFS~\cite{megafs} & {\color[gray]{0.7}0.5905} & {\color[gray]{0.7}0.3802} & 0.6112 & 0.3869 & 0.3499 & \underline{0.4221} & 0.2057 & 0.5408 & 0.0377 & 0.2455 & 0.1607 & 0.4066\tabularnewline
InfoSwap~\cite{infoswap} & {\color[gray]{0.7}0.4308} & {\color[gray]{0.7}0.3056} & 0.4708 & 0.3223 & \textbf{0.2962} & \textbf{0.3963} & 0.2007 & 0.5719 & 0.0192 & 0.1978 & 0.1748 & 0.4250\tabularnewline
FSLSD~\cite{fslsd} & {\color[gray]{0.7}0.7952} & {\color[gray]{0.7}0.4954} & 0.7853 & 0.4623& 0.3706 & 0.4639 & 0.2023 & 0.5385 & 0.0231 & 0.2407 & 0.2598 & 0.5405 \tabularnewline
\midrule
Ours & \textbf{0.4630} & \textbf{0.3987} & {\color[gray]{0.7}0.2574} & {\color[gray]{0.7}0.2316} & 0.3542 & 0.5107 & \underline{0.1537} & \underline{0.4665} & \textbf{0.0125} & \textbf{0.1395} & \textbf{0.1301} & \textbf{0.3514}
\tabularnewline
\bottomrule
\end{tabular}
}} 
\caption{\textbf{Comparison with state-of-the-art methods.} \textbf{Bold} and \underline{underlined} values correspond to the best and the second-best values, respectively. {\color[gray]{0.7}Gray} values are excluded from the evaluation because of the use of the same encoders in their training. $^*$ denotes officially released generated videos. Our method outperforms previous state-of-the-arts in pose and gaze, and achieves the second best results in expression.
}

\label{tab:ff_fid}
\end{table*}

\section{Experiment}
We validate the effectiveness of our method through extensive comparisons with previous methods, ablations and analyses.
The results demonstrate that BlendFace improves the fidelity of identity similarity and attribute preservation compared with previous models.

\subsection{Implementation Detail}
\noindent \textbf{Pretraining of BlendFace.} 
We adopt MS-Celeb-1M~\cite{ms1m} dataset to train BlendFace.
The batch size and the number of epochs are set to $1024$ and $20$, respectively.
We train our encoder on the loss in Eq.~\ref{loss_pretraining} for 20 epochs using SGD optimizer with learning rate $0.1$. We set the probability $p$ of replacing attributes to $0.5$.

\vskip.5\baselineskip
\noindent \textbf{Training of face-swapping model.} 
We adopt VGGFace2~\cite{vggface2} dataset to train our face-swapping model. 
We align and crop the images following FFHQ~\cite{stylegan} preprocessing. 
We use ADAM~\cite{adam} optimizer with $\beta_{1}=0$, $\beta_{2}=0.999$, and $lr=0.0004$ for our generator and discriminator.
We train our model for $300$k iterations with a batch size of 32.
The coefficients of the total loss in Eq.~\ref{total_loss} are empirically set to $\lambda_{1}=10$, $\lambda_{2}=5$, and $\lambda_{3}=5$.

\subsection{Experiment Setup}
\label{sec:main_result}
\noindent \textbf{Setup.} Following the conventional evaluation protocol, we evaluate face-swapping models on FaceForensics++ (FF++)~\cite{ffpp} dataset which includes 1000 real videos and 1000 generated videos each of Deepfakes~\cite{deepfake-faceswap}, Face2Face~\cite{face2face}, FaceSwap~\cite{faceswap}, NeuralTextures~\cite{neuraltextures}, and FaceShifter~\cite{faceshifter}. 
We follow the setting of pairs of source and target defined by the original FF++ dataset. 
And we use the same source frames as in HifiFace~\cite{hififace}.

\vskip.5\baselineskip
\noindent \textbf{Metric.} To evaluate the fidelity of generated images, we consider six metrics: ArcFace (Arc) ~\cite{arcface}, BlendFace (Blend), face shape~\cite{deep3dfacerecon}, expression (Expr)~\cite{deep3dfacerecon}, head pose~\cite{deep3dfacerecon}, and gaze~\cite{gaze}. 
We measure the distances between source and swapped images for identity metrics, \ie ArcFace, BlendFace, and shape, and between target and swapped images for attribute metrics, \ie, expression, pose, and gaze. 
We calculate the cosine distances of extracted feature vectors for ArcFace and BlendFace, L1 distances of predicted 3DMM parameters for shape, expression, and pose, and L1 distances of predicted Euler angles for gaze.
We further compute the relative distances~\cite{smoothswap} that consider both source and target in all the metrics, which are denoted as ``-R''.

\subsection{Comparison with Previous Methods}
\label{sec:comparison_sota}
\vskip.5\baselineskip
\noindent \textbf{Baselines.} We compare our method with publicly available state-of-the-art models, \ie,  FSGAN~\cite{fsgan}, SimSwap~\cite{simswap}, MegaFS~\cite{megafs}, InfoSwap~\cite{infoswap}, and FSLSD~\cite{fslsd}. We also adopt generated videos of Deepfakes~\cite{deepfake-faceswap}, FaceSwap~\cite{faceswap}, and FaceShifter~\cite{faceshifter} from FF++ dataset and HifiFace~\cite{hififace} from the official project page~\cite{hififace_page}. 

\begin{figure*}[t]
    \centering
    \includegraphics[width=0.99\linewidth]{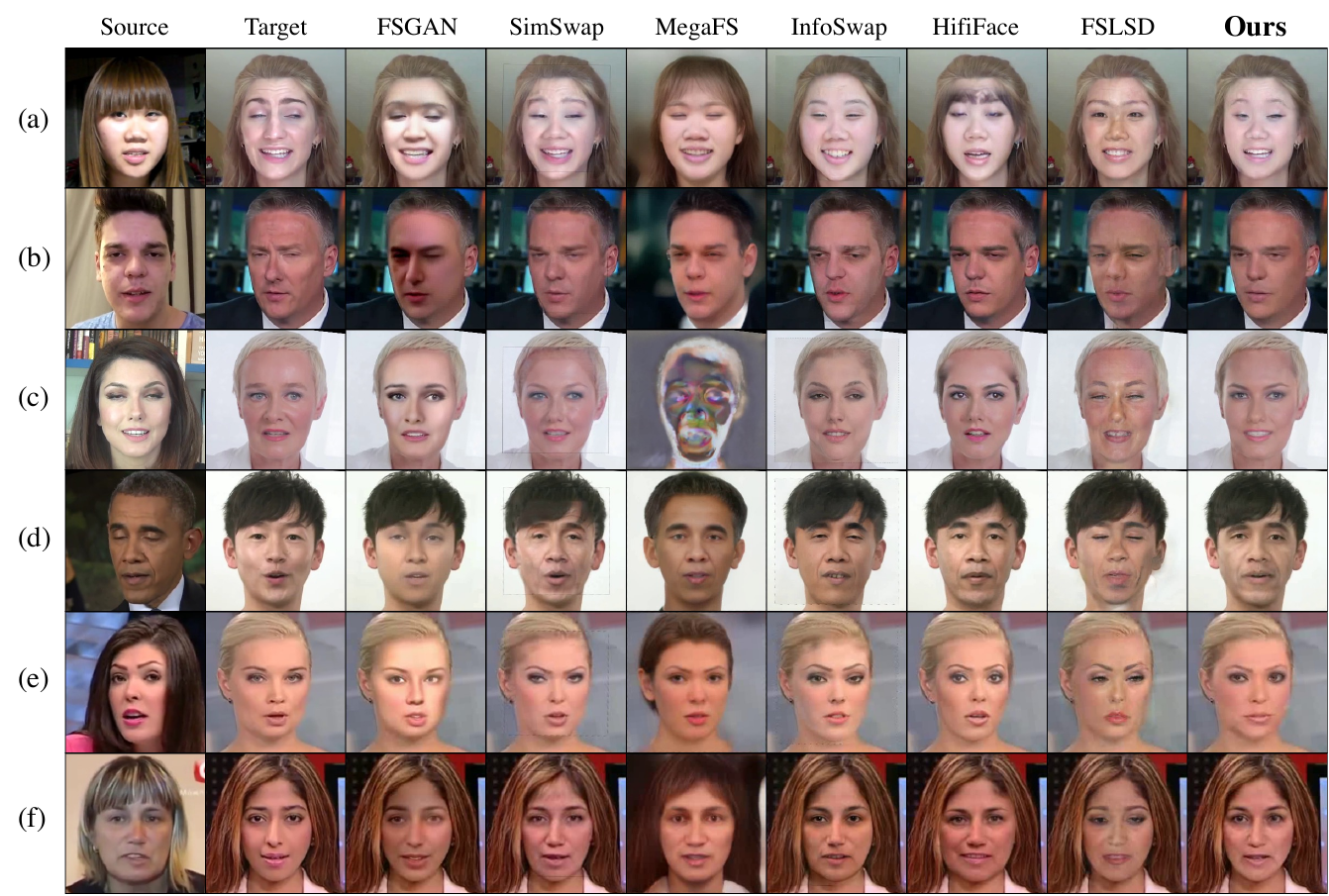}
    \caption{\textbf{Qualitative comparison on FF++.} Our model performs consistent face-swapping while preserving target attributes, \eg, hairstyles, expression, pose, and gaze. More results are included in the supplementary material.}
    \label{fig:ff}
\end{figure*}

\vskip.5\baselineskip
\noindent \textbf{Result.}
The quantitative result is presented in Table~\ref{tab:ff_fid}.
Our model outperforms previous methods in terms of absolute and relative metrics of pose and gaze, and achieves the second-best result in absolute and relative metrics of expression.
Although it is difficult to compare our model with previous methods in identity similarity using face recognition models because evaluations using the same encoder as in training tends to be overestimated, our model outperforms Deepfakes~\cite{deepfake-faceswap}, FaceSwap~\cite{faceswap}, and FSGAN~\cite{fsgan} in ArcFace at least.
We also show some generated images in Fig.~\ref{fig:ff}. 
We omit examples of Deepfakes, FaceSwap, and FaceShifter because these results are generated by different source frames. 
We can see that previous approaches suffer from attribute leakages, particularly in changing hairstyles, (\eg, (a) and (d)) and head shapes, (\eg, (c) and (e)), whereas our method succeeds in generating robust results qualitatively. 
In addition, we observe that the StyleGAN2-based methods, \ie, MegaFS~\cite{megafs} and FSLSD~\cite{fslsd} are vulnerable to unseen domain because the models sometimes fail in the inversion of source identities. 
We include more generated results in the supplementary material due to the space limitations.
Because our method limits face shape changes to maintain consistency between the inside and outside of faces, the similarities of face shapes between generated images and source images are slightly lower than those of other methods, which is one of the limitations of our approach discussed in Sec.~\ref{sec:limitation}.


\subsection{Ablation and Analysis}
\label{sec:ablation}

\begin{table}[t]
    
    \centering
    \begin{adjustbox}{width=0.475\textwidth}
    \begin{tabular}{cccccc} \toprule
      \multicolumn{2}{c}{Setting} & \multicolumn{4}{c}{Distance}\\
      \cmidrule(lr){1-2} \cmidrule(lr){3-6}
      Source&Loss & Arc &Blend & Pose & Gaze\\
      \midrule
      ArcFace&ArcFace& \textbf{0.3314} & 0.2918  & 0.0139 & 0.1450 \\
      BlendFace&ArcFace& 0.4254 & 0.2730  &  \underline{0.0128}  & \underline{0.1376} \\
      ArcFace&BlendFace&\underline{0.4123} & \underline{0.2700}  & 0.0131  & 0.1377 \\
      BlendFace&BlendFace& 0.4630 & \textbf{0.2574}  &  \textbf{0.0125} & \textbf{0.1301} \\
      \bottomrule
    \end{tabular}
    \end{adjustbox}
    \caption{\textbf{Choices of source encoder and loss.} Our method achieves the best results in distances of BlendFace, pose, and gaze.
    }
    \label{tab:ablation_choice}
\end{table}

\begin{figure}[t]
    \centering
    \includegraphics[width=0.99\linewidth]{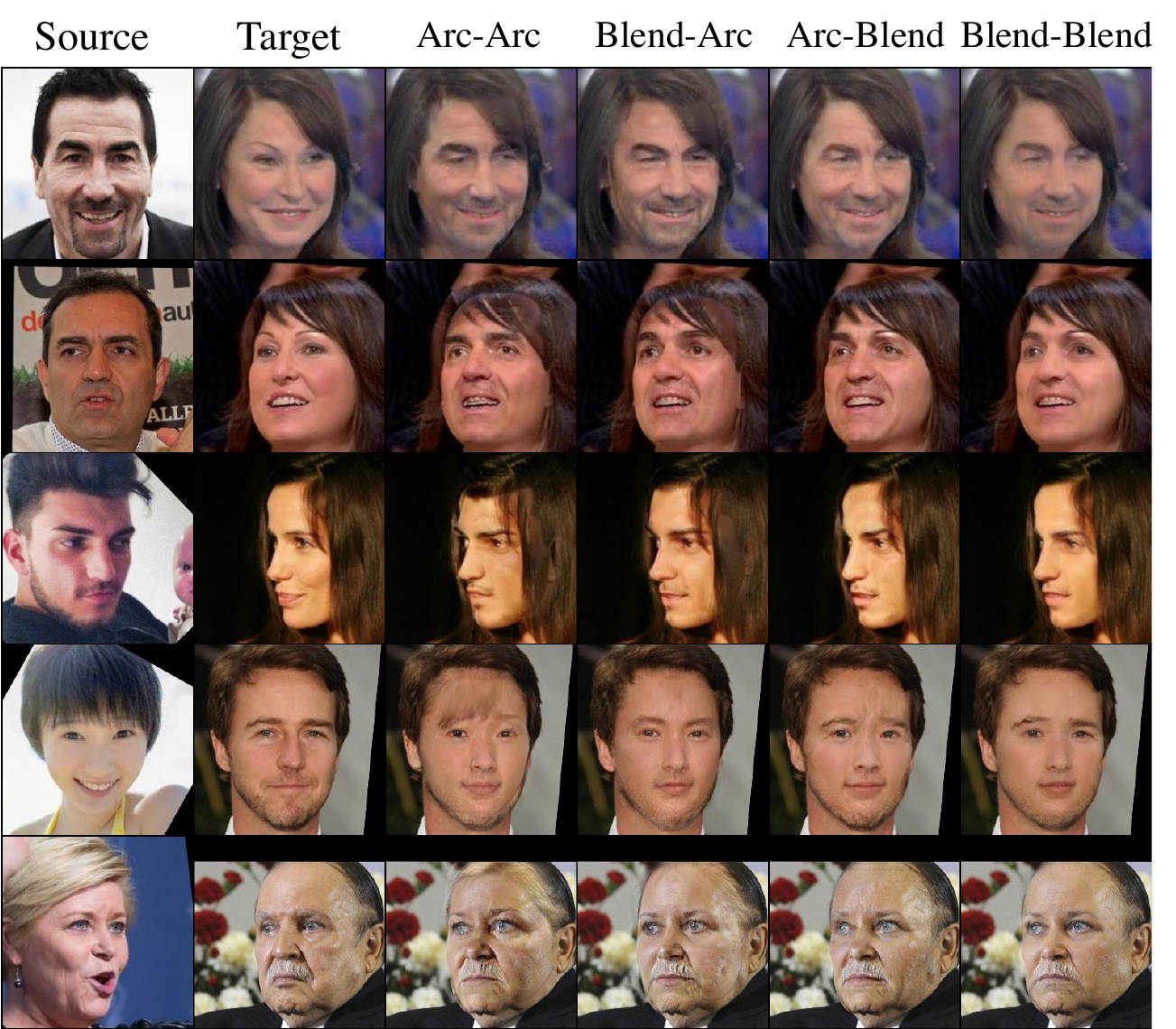}
    \caption{\textbf{Choices of encoder and loss.} The notation ``A-B'' indicates using encoders A and B for source feature extraction and loss computation, respectively. Our model produces reasonable results in the identity similarity and attribute preservation. Best viewed in zoom.}
    \label{fig:ablation_choice}
\end{figure}

\noindent \textbf{Choices of source encoder and loss.} 
We found that the performance of face-swapping models strongly rely on choices of identity encoders to extract source features and compute the identity loss. We train four face-swapping models with ArcFace and BlendFace set to source encoder or loss. Then we evaluate these models in the same protocol as Sec.~\ref{sec:comparison_sota}. Note that our model in Sec.~\ref{sec:comparison_sota} sets BlendFace both to identity encoder and loss.
The result is given in Table~\ref{tab:ablation_choice}. 
It can be observed that using ArcFace both in the source encoder and loss computation brings the worst results in distances of BlendFace, expression, pose, and gaze, which implies the identity-attribute entanglement of ArcFace. 
Our method meanwhile achieves the best preservation in pose and gaze though worst identity distance on ArcFace because ArcFace tends to underestimate well-disentangled swapped results.
We also show the generated images in Fig.~\ref{fig:ablation_choice}. 
We can observe computing the identity loss with ArcFace produces inconsistent blending boundaries and attributes because of the entanglement of identities and attributes, while BlendFace produces consistent results.
These results clearly indicate that our BlendFace outperforms ArcFace quantitatively and qualitatively in face-swapping.

\begin{figure}[t]
    \centering
    \begin{minipage}[b]{0.49\linewidth}
    \centering
    \includegraphics[width=1.0\linewidth]{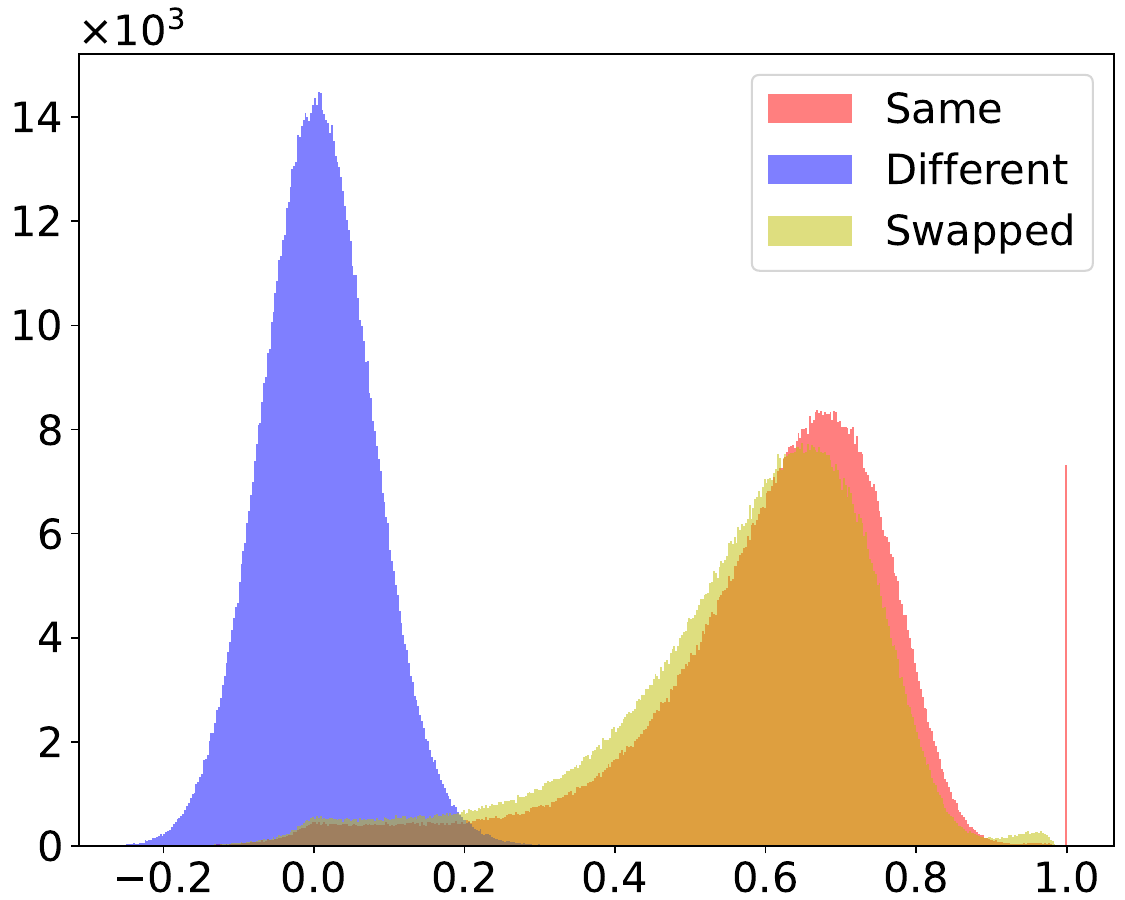}
    \subcaption{$p=0.25$}
    \label{hist_p025}
    \end{minipage}
    \begin{minipage}[b]{0.49\linewidth}
    \centering
    \includegraphics[width=1.0\linewidth]{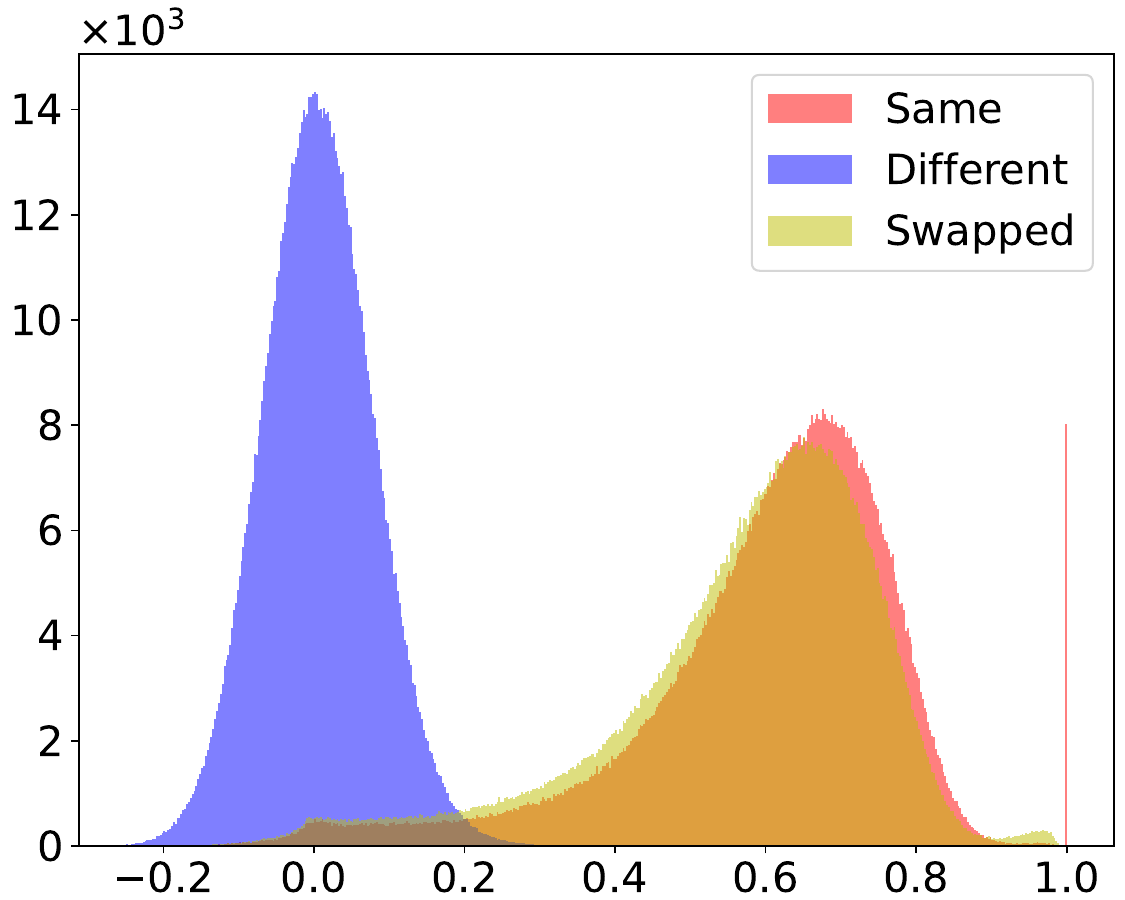}
    \subcaption{$p=0.5$}
    \label{hist_p050}
    \end{minipage}\\
    \begin{minipage}[b]{0.49\linewidth}
    \centering
    \includegraphics[width=1.0\linewidth]{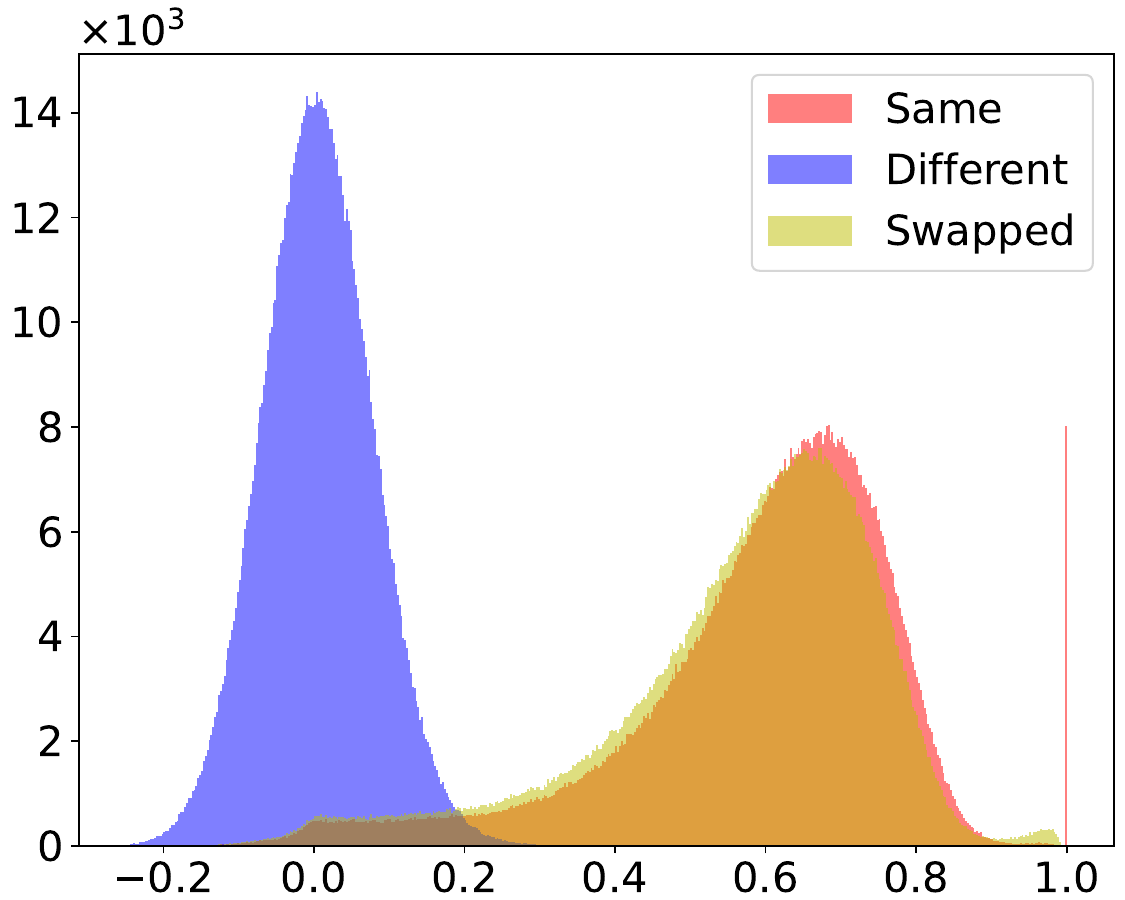}
    \subcaption{$p=0.75$}
    \label{hist_p075}
    \end{minipage}
    \begin{minipage}[b]{0.49\linewidth}
    \centering
    \includegraphics[width=1.0\linewidth]{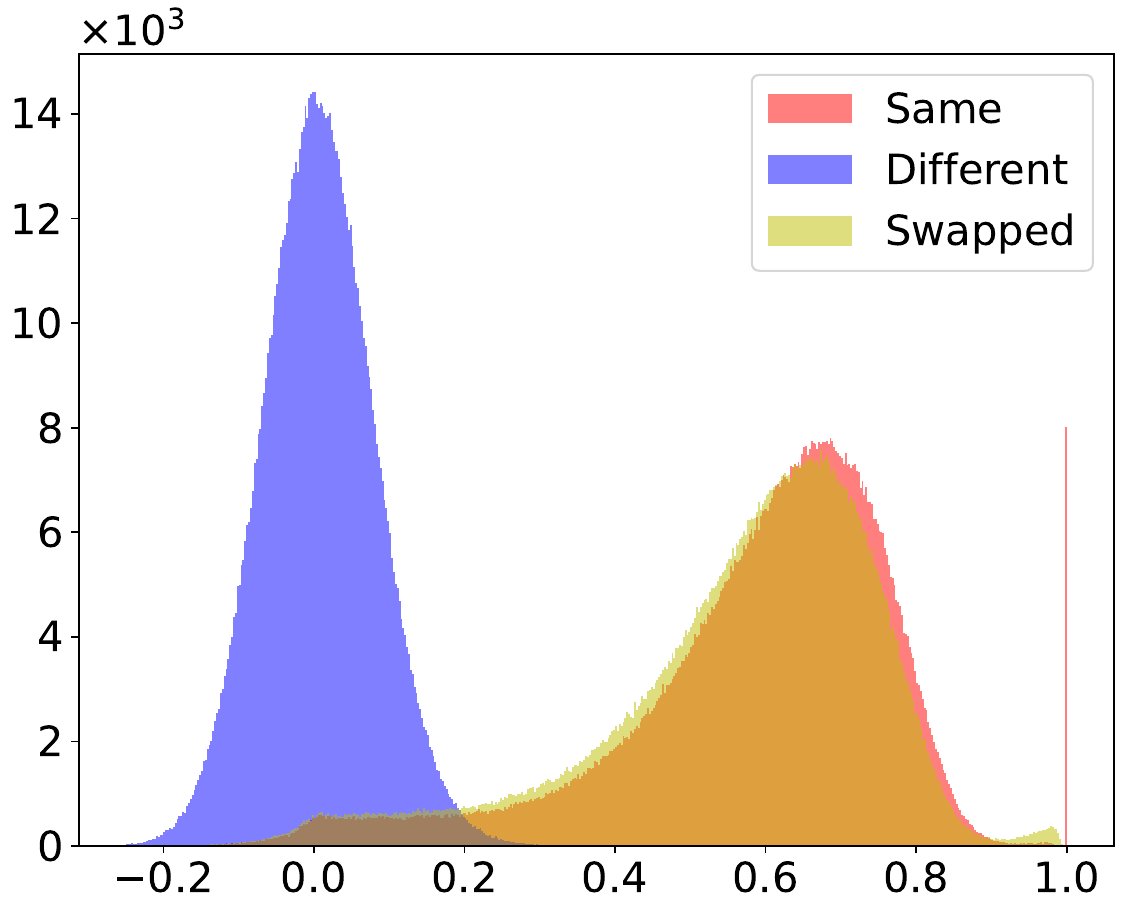}
    \subcaption{$p=1.0$}
    \label{hist_p100}
    \end{minipage}
    \caption{\textbf{Effect of the ratio of synthetic faces.} Training identity encoders with swapped faces brings the distributions of swapped faces and that of real positive faces closer together.}
    \label{fig:ablation_ratio_hist}
\end{figure}

\vskip.5\baselineskip
\noindent \textbf{Probability of replacing attributes.} 
We examine the effect of the probability $p$ of replacing attributes in pretraining of the identity encoder. 
We additionally train our encoders with $p=0.25$, $0.75$, and $1.00$, then we conduct the same experiment as Sec.~\ref{sec:preliminary} and plot the distributions in Fig~\ref{fig:ablation_ratio_hist}. 
It can be observed that training identity encoders with swapped faces bridges the gap between the similarity distributions of actual positive samples denoted as ``Same'' and swapped faces denoted as ``Swapped'' in the figure, even when $p=0.25$. 
Also, increasing $p$ brings these distributions closer.
We again emphasize that ArcFace~\cite{arcface}, which corresponds to the case of $p=0.0$, underestimates swapped faces as shown in Fig.~\ref{preliminary}. 
This result supports that our pre-training approach removes attribute biases from face recognition models.

\begin{table}[t]
    
    \centering
    \begin{adjustbox}{width=0.41\textwidth}
    \begin{tabular}{lccc} \toprule
      Model&LFW&CFP-FP&AgeDB\\
      \midrule
      ArcFace~\cite{arcface} & 0.9978 & \underline{0.9900} & \textbf{0.9838}\\
      SynFace~\cite{synface} & 0.9763 & 0.8731& 0.8243 \\
      BlendFace ($p=0.25$) & \underline{0.9983} & \textbf{0.9904} & 0.9815 \\
      BlendFace ($p=0.50$) & \textbf{0.9984} &0.9897 &0.9813\\
      BlendFace ($p=0.75$) &0.9983&0.9891& \underline{0.9828}\\
      BlendFace ($p=1.00$) & 0.9977 & 0.9886 & 0.9800 \\
      \bottomrule
    \end{tabular}
    \end{adjustbox}
    \caption{\textbf{Face verification result.} Our encoders keep the ability of face verification on real face datasets and outperforms SynFace, which support our encoders perform properly as identity guidance in training face-swapping models.}
    \label{tab:retrieval}
\end{table}

\vskip.5\baselineskip
\noindent \textbf{Face verification on real face datasets.}
We validate our encoders in the task of face verification.
We adopt well-known benchmarks including LFW~\cite{lfw}, CFP-FP~\cite{cfpfp}, and AgeDB~\cite{agedb}.
We compare our BlendFace with ArcFace and SynFace~\cite{synface}, a variant of ArcFace trained on GAN-synthesized images.
We adopt the \textit{unrestricted with labelled outside
data} protocol to evaluate models, following the convention in the research field of face recognition (\eg, \cite{arcface}).
We give the results in Table~\ref{tab:retrieval}.
We observe that our encoders retain the ability of face verification despite the slight performance degradation. This is because the removed features, \eg, for hairstyles, face-shapes, and colors, are useful in verifying real faces.

\begin{figure}[t]
    \centering
    \includegraphics[width=0.99\linewidth]{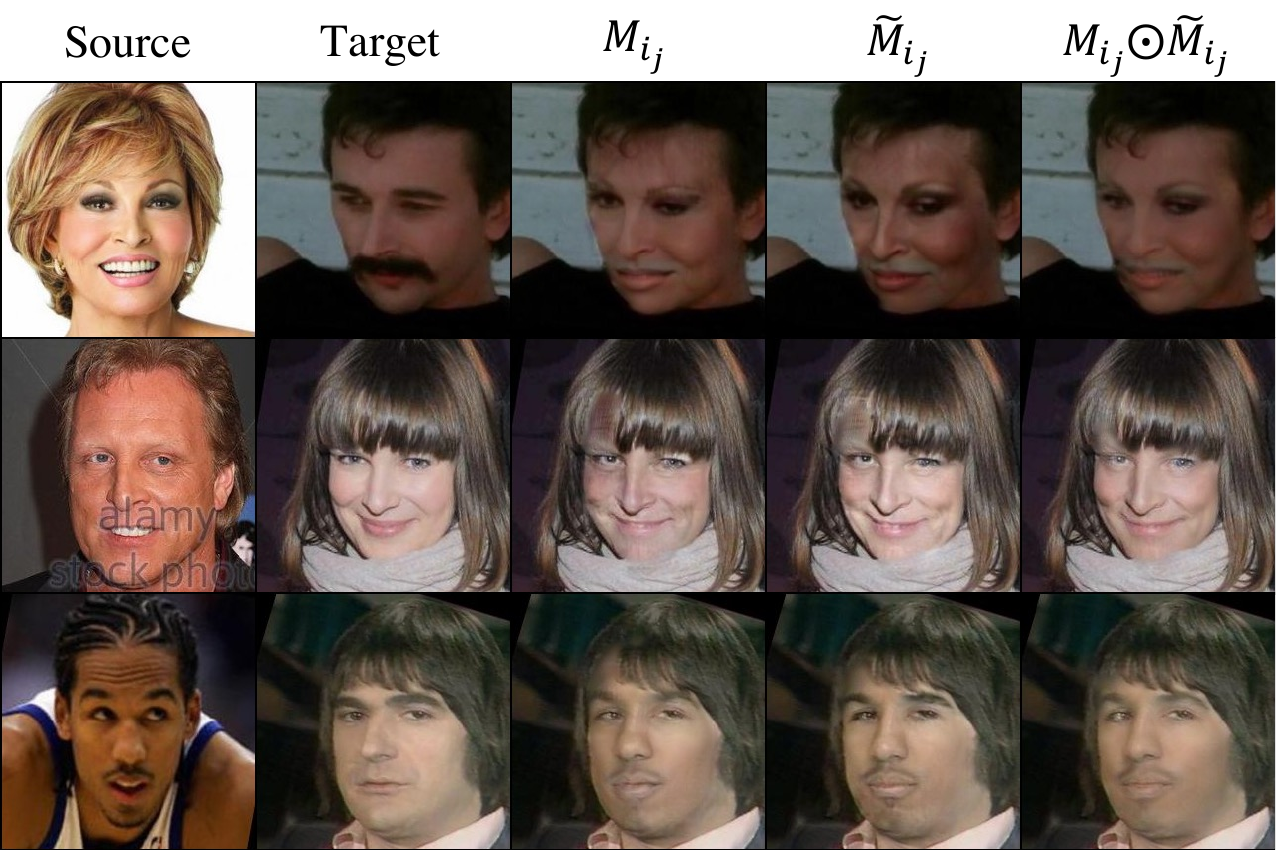}
    \caption{\textbf{Effect of blending masks.} Using intersection mask ${M}_{i_j} \odot \tilde{M}_{i_j}$ contributes to the consistency of generated images.}
    \label{fig:ablation_mask}
\end{figure}

\vskip.5\baselineskip
\noindent \textbf{Blending mask.} 
We examine the effect of blending masks in pretraining of BlendFace. 
As described in Sec.~\ref{sec:preliminary}, we use a mask $\hat{M}_{i_j}$ generated from the intersection $M_{i_j} \odot \tilde{M}_{i_j}$ to blend source image $X_{i_j}$ and target image $\tilde{X}_{i_j}$ during pretraining.
We here train BlendFace with $M_{i_j}$ or $\tilde{M}_{i_j}$ instead of ${M}_{i_j} \odot \tilde{M}_{i_j}$. Then we train face-swapping models with these encoders.
As shown in Fig.~\ref{fig:ablation_mask}, we found the model with ${M}_{i_j} \odot \tilde{M}_{i_j}$ produces more consistent results than ${M}_{i_j}$ or $\tilde{M}_{i_j}$. 
This is because blending $X_{i_j}$ and $\tilde{X}_{i_j}$ with $M_{i_j}$ or $\tilde{M}_{i_j}$ produces artifacts in the blended image $\hat{X}_{i_j}$ during pretraining, which harms face-swapping models.

\begin{figure}[t]
    \centering
    \includegraphics[width=0.99\linewidth]{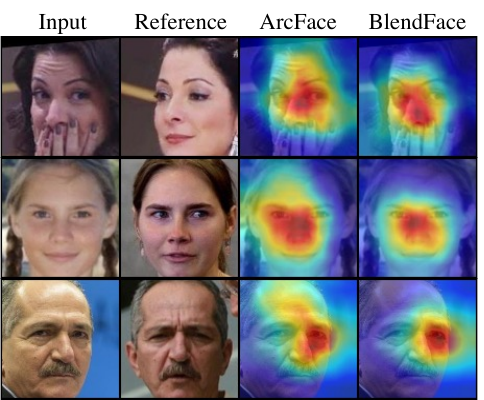}
    \caption{\textbf{Saliency maps of ArcFace and BlendFace.} ArcFace pays attention not only to inner faces but also to outer faces while BlendFace does only to inner faces.}
    \label{fig:saliency}
\end{figure}

\vskip.5\baselineskip
\noindent \textbf{Saliency Map.} 
To explore the effectiveness of BlendFace, we visualize the saliency maps of ArcFace and BlendFace. Inspired by sliding occlusions~\cite{occlusion}, we measure the averaged sensitivity of identity similarities between the occluded input images and the reference image over multiple mask sizes $\{16,24,32,40,48,56\}$. 
As shown in Fig.~\ref{fig:saliency}, ArcFace focuses on both the inner and outer faces while our encoder properly does only on inner faces.
The result supports that our BlendFace can swap only inner faces without the undesired attribute transfer of outer faces.

\section{Limitations}
\label{sec:limitation}
Our novel identity encoder BlendFace provides disentangled identity features that are beneficial face-swapping and other face-related tasks; however, we notice some limitations of our method. 
First, our model can hardly change face shapes because we limit the region where identities are swapped to improve spatial consistencies between inside and outside of faces. Therefore when source and target images that have to different face shapes are input into our model, the generated image looks like the source subject in terms of the appearance but may not in terms of the face shape.
Second, similar to previous methods, our method sometimes fail to preserve hard occlusions such as hands because of the lack of training samples of extreme scenes. It can be improved by incorporating HEAR-Net~\cite{faceshifter} into our model.

\section{Conclusion} 
In this paper, we present BlendFace, a well-disentangled identity encoder for more consistent face-swapping. The key observation behind BlendFace is that traditional face recognition models trained on real face datasets have biases in some attributes, \eg, hairstyles and face-shapes, which leads inconsistent results in face-swapping. To tackle this problem, we train a face recognition model with blended images as pseudo-positive samples that have swapped attributes so that the encoder focuses only on inner faces, which improves the disentanglement of identity and attribute for face-swapping.
The comparison with previous methods on FaceForensics++ dataset demonstrates our method achieves new state-of-art results especially in preserving target attributes, keeping the visual consistency.
Also our extensive analyses provide the advantages of BlendFace for subsequent face-related research.

\vskip.5\baselineskip
\noindent\textbf{Potential Negative Societal Impacts.}
Face Swapping models are at risk of abuse, \textit{e.g.}, deceiving face verification systems and synthesizing political speeches, that are known as ``deepfake''. Therefore the vision community has been working on digital face forensics, which leads to so many promising deepfake detection approaches~\cite{pcl,fakespotter,realforensics,idreveal,headpose,eyeblink,mesonet,twobranch,ftcn,lipforensics,sbi,facexray} and a wide variety of benchmarks~\cite{ffpp,celebdf,dfdc,dfdcp,ffiw,kodf,wilddeepfake,deeperforensics,fakeavceleb,dftimit}. The risk can be mitigated by proactive detection methods~\cite{yu2021artificial,yang2021faceguard} and by strictly gating the release of our model only for research purpose. In addition, we will release the benchmark dataset of our model on FF++ for future studies of face forensics.

{\clearpage\small
\bibliographystyle{ieee_fullname}
\bibliography{egbib}
}
\clearpage\appendix

\section{Architecture}
\begin{figure}[t]
    \centering
    \includegraphics[width=1.0\linewidth]{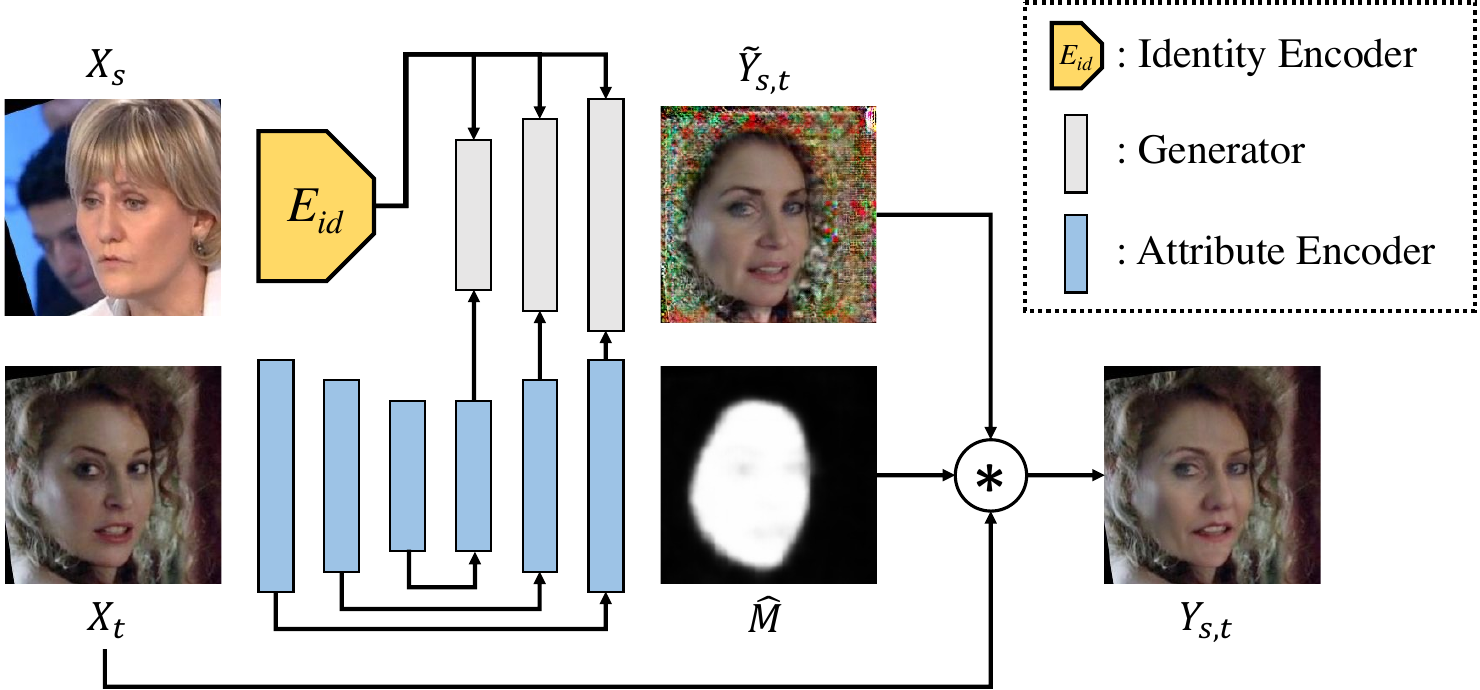}
    \caption{\textbf{Architecture of our model.} We simply add a convolution and sigmoid layer to the last layer of the original attribute encoder from AEI-Net~\cite{faceshifter}.}
    \label{fig:architecture}
\end{figure}

Here, we detail the architecture of our face-swapping model.
As described in the main paper, our architecture is based on AEI-Net ~\cite{faceshifter} but we incorporate a mask predictor into the model inspired by previous studies~\cite{hififace,styleswap,rafswap}. 
We illustrate the architecute of our model in Fig.~\ref{fig:architecture}.
We simply add a convolution layer and sigmoid layer to the last layer of the attribute encoder to predict blending masks $\hat{M}$. 
We blend the foreground face image $\tilde{Y}_{s,t}$ and target image $X_t$ using the predicted $\hat{M}$  as follows:
\begin{equation}
\label{eq:output_blending}
    Y_{s,t}=\tilde{Y}_{s,t} \odot \hat{M} + X_t \odot (1-\hat{M}).
\end{equation}
In our model, we assume the blending masks for the same target images should be the same independently of source images.
Note that losses are computed on the blended result $Y_{s,t}$; therefore, the intermediate generated face $\tilde{Y}_{s,t}$ is noisy outside of the face.

\begin{figure}[h]
    \centering
    \includegraphics[width=1.0\linewidth]{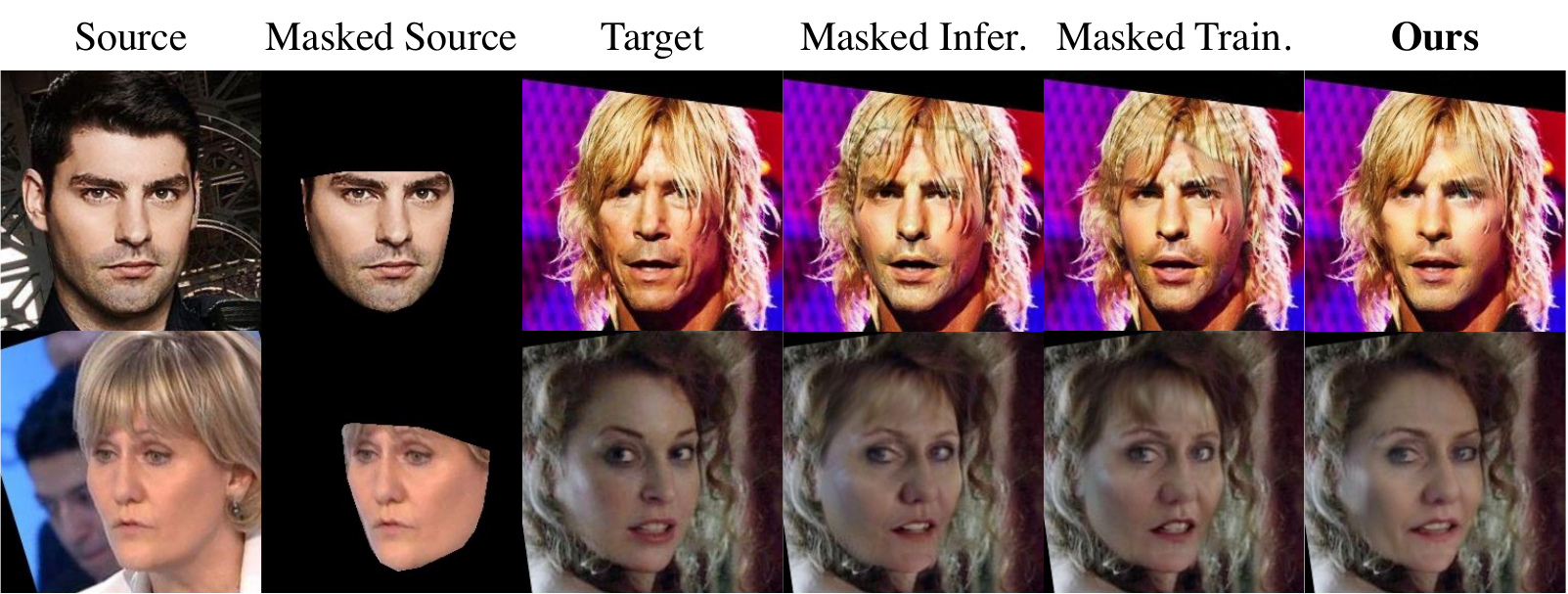}
    \caption{\textbf{Comparison with additional baselines.}}
    \label{industry}
\end{figure}

\section{Comparison with Additional Baselines} 
We compare our model with two additional baselines in Fig.~\ref{industry}:
1) We input masked source images into a pretrained face-swapping model that is the same model as Arc-Arc in Fig.~\textcolor{red}{6}, denoted as \textit{Masked Infer.}. 
To only include the face area below the eyebrow and between ears, we generate the masks by computing the convex hull of 68 facial landmarks by FAN~\cite{fan}.
2) We train Arc-Arc model from scratch with masked source images, denoted as \textit{Masked Train.}.
As can be seen, the models still suffer from the attribute inconsistency. 
Therefore, we can conclude our method is a unique solution for the attribute leakage problem.

\begin{figure}[h]
    \centering
    \begin{minipage}[b]{0.32\linewidth}
    \centering
    \includegraphics[width=1.0\linewidth]{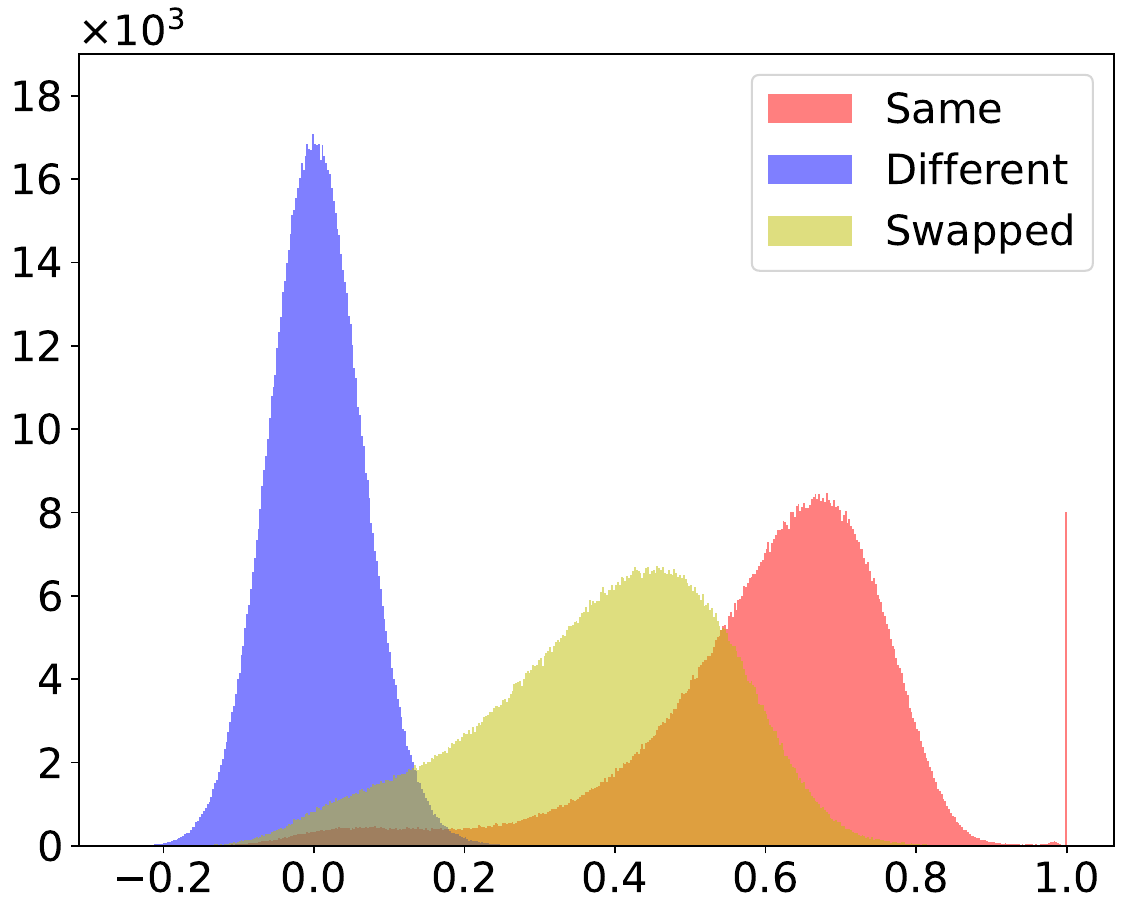}
    \subcaption{{\scriptsize Cur-R100-MS1M}}
    \end{minipage}
    \begin{minipage}[b]{0.32\linewidth}
    \centering
    \includegraphics[width=1.0\linewidth]{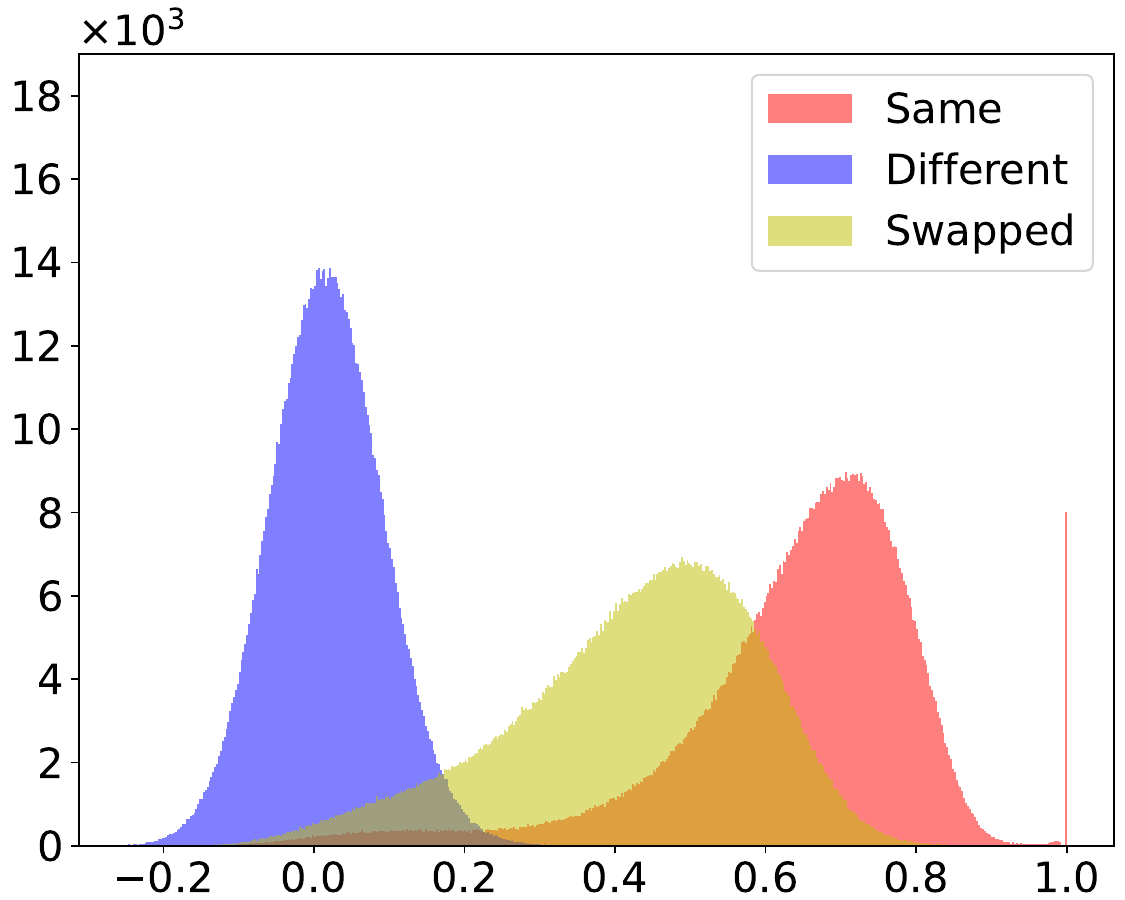}
    \subcaption{{\scriptsize QMag-R100-MS1M}}
    \end{minipage}
    \begin{minipage}[b]{0.32\linewidth}
    \centering
    \includegraphics[width=1.0\linewidth]{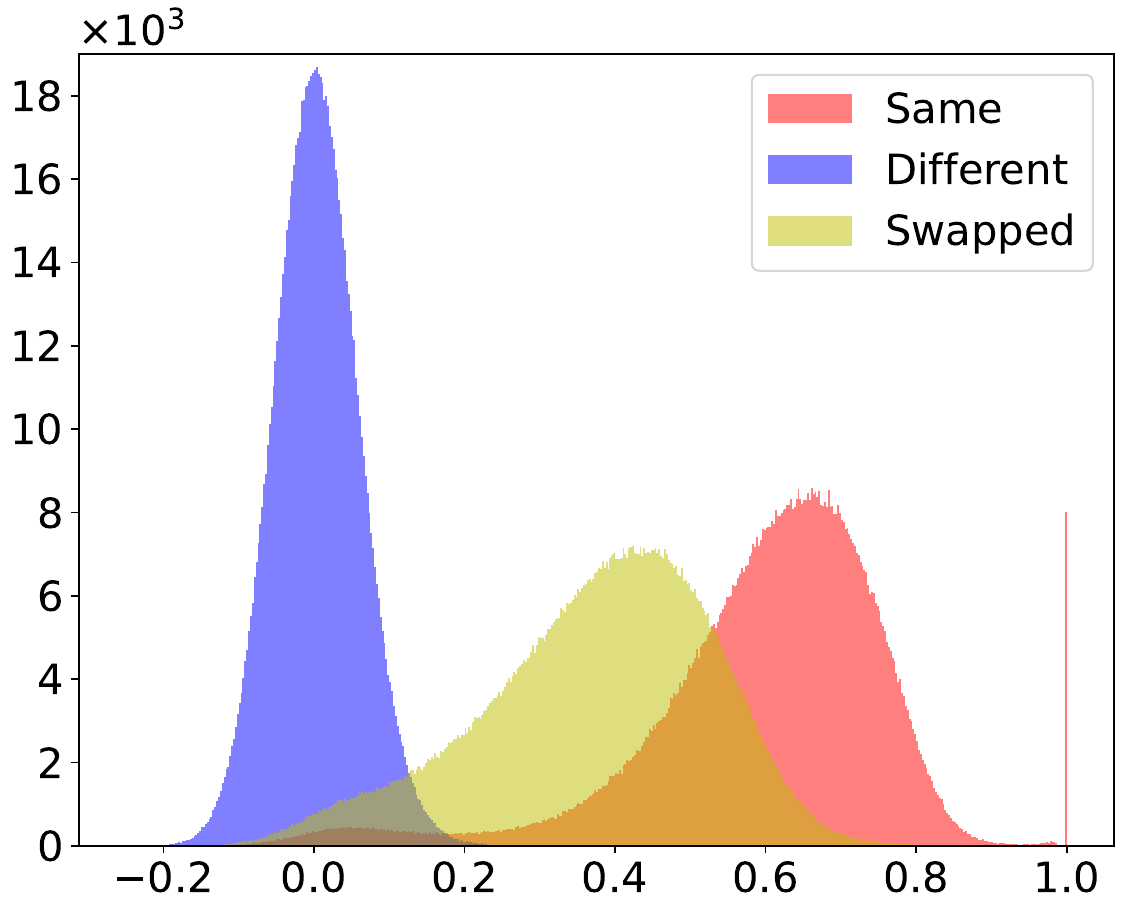}
    \subcaption{{\scriptsize Ada-R100-WF12M}}
    \end{minipage}\\

    \begin{minipage}[b]{0.32\linewidth}
    \centering
    \includegraphics[width=1.0\linewidth]{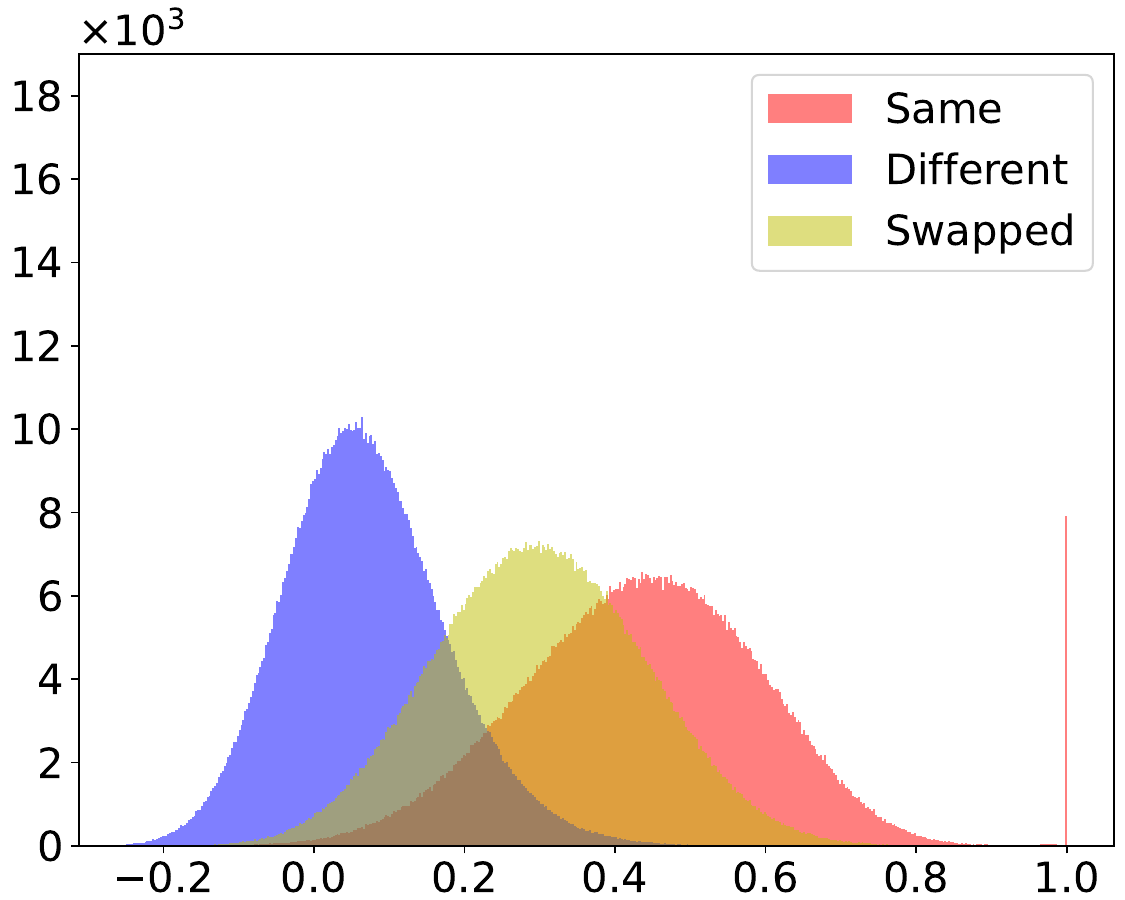}
    \subcaption{{\scriptsize Arc-R50-Syn}}
    \end{minipage}
    \begin{minipage}[b]{0.32\linewidth}
    \centering
    \includegraphics[width=1.0\linewidth]{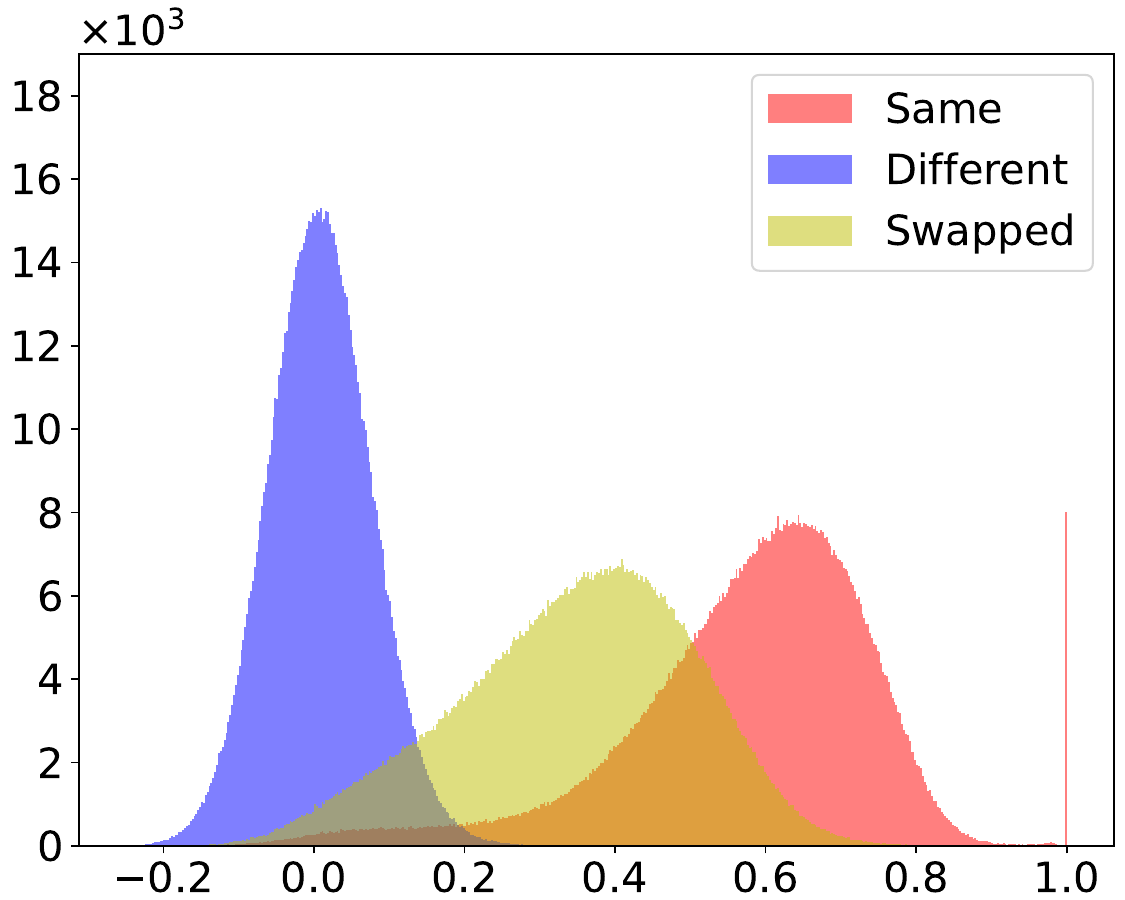}
    \subcaption{{\scriptsize Arc-ViT-MS1M}}
    \end{minipage}
    \begin{minipage}[b]{0.32\linewidth}
    \centering
    \includegraphics[width=1.0\linewidth]{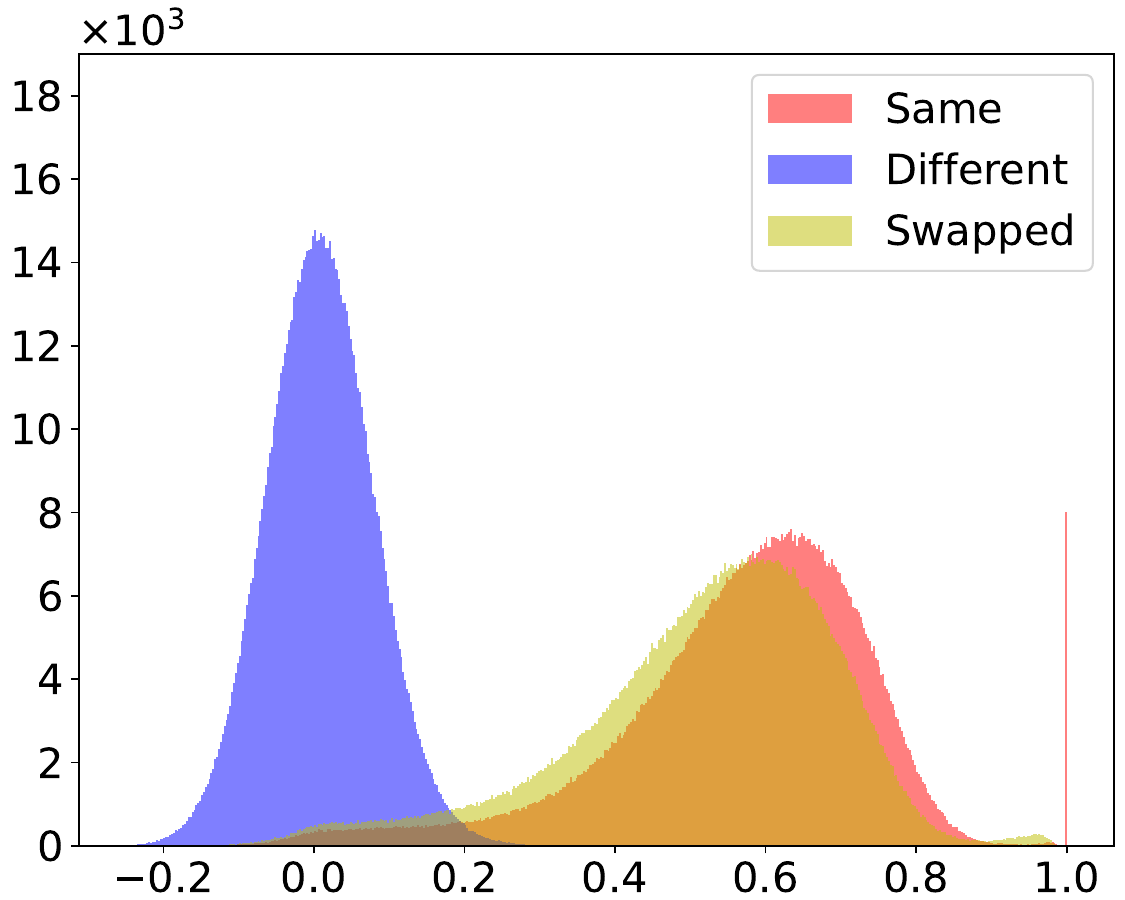}
    \subcaption{{\scriptsize \textbf{Blend}-ViT-MS1M}}
    \end{minipage}
    \caption{\textbf{Different models and dataset.}}
    \label{fig:different_models}
\end{figure}

\section{Attribute Biases on Different Face Recognition Models} 
We visualize the similarity distributions of different face recognition models in Fig.~\ref{fig:different_models}: (a)~CurricularFace (Cur)~\cite{curricularface}, (b)~QMagFace (QMag)~\cite{qmagface}, (c)~AdaFace (Ada)~\cite{adaface} on a lager dataset WebFace12M (WF12M)~\cite{webface260m}, (d)~ArcFace (Arc) on GAN-generated face images (Syn)~\cite{synface}, and (e) ArcFace with VisionTransformer (ViT)~\cite{vit}.
We can see that the attribute leakage problem still exists in these various face recognition models.
Additionally, we train ViT backbone using our BlendFace pretraining.
As shown in Fig.~\ref{fig:different_models}~(f), our training strategy works well even on ViT, which proves the generality of our method.

\section{More Comparisons on FF++}
\begin{figure*}[t]
    \centering
    \includegraphics[width=0.925\linewidth]{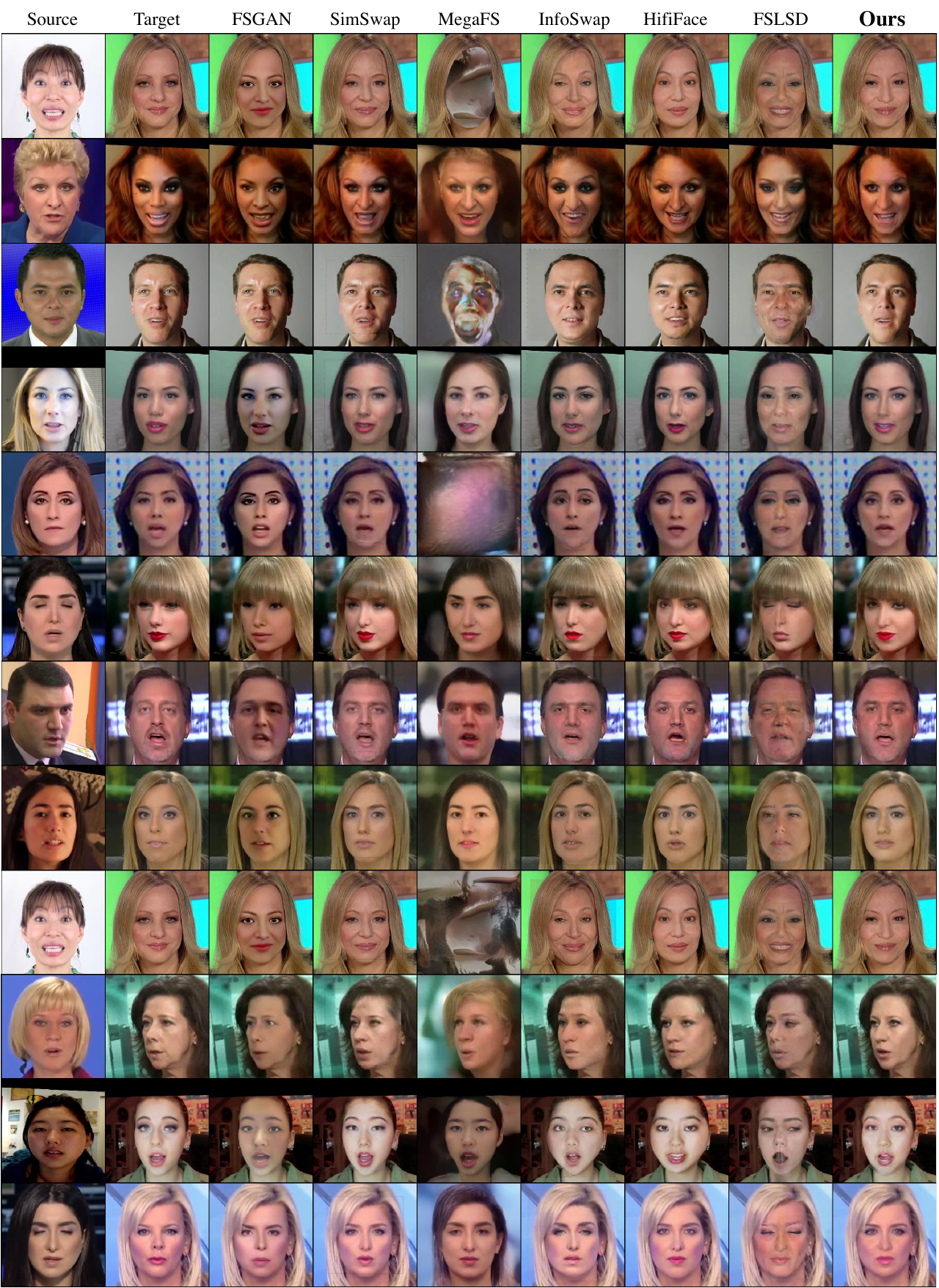}
    \caption{\textbf{More comparisons on FF++.}}
    \label{fig:more_result}
\end{figure*}

We show more qualitative comparisons on FaceForensics++~\cite{ffpp} in Fig.~\ref{fig:more_result}.
Our model performs consistent face-swapping for a range of source and target images compared to previous methods~\cite{simswap,infoswap,megafs,fslsd,fsgan,hififace}.

\end{document}